\documentclass{article}

\usepackage{array}
\usepackage{microtype}
\usepackage{graphicx}
\usepackage{diagbox}
\usepackage{bbold}
\usepackage{mathtools}
\usepackage{multirow}
\usepackage{amsmath}
\usepackage{subcaption}
\usepackage{color}
\usepackage{booktabs} 
\usepackage{amssymb}
\usepackage{pifont}
\newcommand{\cmark}{\ding{51}}%
\usepackage{hyperref}

\usepackage[accepted]{icml2024}

\usepackage{amsmath}
\usepackage{amssymb}
\usepackage{mathtools}
\usepackage{amsthm}

\usepackage[capitalize,noabbrev]{cleveref}

\theoremstyle{plain}
\newtheorem{theorem}{Theorem}[section]
\newtheorem{proposition}[theorem]{Proposition}
\newtheorem{lemma}[theorem]{Lemma}

\theoremstyle{definition}
\newtheorem{definition}[theorem]{Definition}

\theoremstyle{remark}

\usepackage[textsize=tiny]{todonotes}

\icmltitlerunning{MultiMax: Sparse and Multi-Modal Attention Learning}

\begin{document}

\twocolumn[
\icmltitle{MultiMax: Sparse and Multi-Modal Attention Learning}





\icmlsetsymbol{equal}{*}

\begin{icmlauthorlist}
\icmlauthor{Yuxuan Zhou}{a,b}
\icmlauthor{Mario Fritz}{b}
\icmlauthor{Margret Keuper}{a,c}
\end{icmlauthorlist}
\icmlaffiliation{a}{University of Mannheim, Germany}
\icmlaffiliation{b}{CISPA Helmholtz Center for Information Security, Germany}
\icmlaffiliation{c}{Max Planck Institute for Informatics, Saarland Informatics Campus, Germany}
\icmlcorrespondingauthor{Mario Fritz}{fritz@cispa.de}
\icmlcorrespondingauthor{Margret Keuper}{keuper@mpi-inf.mpg.de}
\icmlkeywords{Machine Learning, ICML}
\vskip 0.3in
]



\printAffiliationsAndNotice{}  

\begin{abstract}
SoftMax is a ubiquitous ingredient
of modern machine learning algorithms. It maps an input vector onto a probability simplex and reweights the input by concentrating the probability mass at large entries. 
Yet, as a smooth approximation to the Argmax function, a significant amount of  probability mass is distributed to other, residual entries, leading to poor interpretability and noise. Although sparsity can be achieved by a family of SoftMax variants, they often require an alternative loss function and do not preserve multi-modality. 
We show that this trade-off between multi-modality and sparsity limits the expressivity of SoftMax as well as its variants. 
We provide a solution to this tension between objectives by proposing a piece-wise differentiable function, termed MultiMax, which adaptively modulates the output distribution according to input entry range. Through comprehensive analysis and evaluation, we show that MultiMax successfully produces a distribution that supresses irrelevant entries while preserving multi-modality, with benefits in image classification, language modeling and machine translation. The code is available at \url{https://github.com/ZhouYuxuanYX/MultiMax}.
\end{abstract}

\section{Introduction}
The SoftMax has remained in wide use in modern machine learning methods and finds its application in a variety of algorithms such as multi-class classification \cite{lecun2015deep, goodfellow2016deep, bishop2006pattern}, attention mechanisms \cite{vaswani2017attention, velivckovic2017graph, bahdanau2014neural, gehring2016convolutional} and reinforcement learning \cite{sutton2018reinforcement, rummery1994line, williams1992simple}. It can be regarded as a differentiable approximation of the Argmax operation and projects the input onto the probability simplex, which allocates most of the probability mass to large entries. From the perspective of optimization, the SoftMax function allows for a reasonable trade-off between exploitation and exploration \cite{white1992role}, i.e., important positions are emphasized while every position has a chance of being explored. This trade-off can be controlled by a scale factor, which is often referred to as temperature.

However, the expressivity of SoftMax is severely limited by the following dilemma: a high temperature leads to over-smoothing and reduces the efficiency of the optimization, whereas a small temperature collapses multi-modality and makes training unstable. In attention layers for example, a small temperature will cause relevant positions except the peak to be overlooked, whereas a large temperature will ``waste" a non-negligible portion of attention  on irrelevant keys. Therefore, temperature is often set to one by default in attention layer. As shown later, such a compromise also results in the recently observed over-smoothing issue in both vision \cite{gong2021improve, wang2022anti} and language \cite{shi2022revisiting} transformers. Moreover, transformer-based Large Language Models are shown to be prone to the interference of irrelevant context \cite{shi2023large, jia2017adversarial}, which is also highly related to the portion of attention on irrelevant tokens \cite{weston2023system}. 

To overcome the issue, previous works have proposed sparse SoftMax alternatives, which allow to completely ignore small entries below a threshold. These sparse SoftMax variants have been studied in diverse contexts, e.g., generative modeling \cite{chen2021evidential}, output activations of multi-class classifiers, and/or attention mechanisms \cite{peters2019sparse, martins2016softmax, 
gupta2021memory}. 

However, such methods often suffer from poor gradient signal, which leads to instability during training. Moreover, the number of non-sparse dimensions is often treated as empirically selected hyperparameter.

\begin{figure*}[t]
\centering
    \begin{subfigure}[t]{0.528\textwidth}
    \centering
    \includegraphics[trim={0cm 0cm 0 0cm},clip, width=0.9\textwidth]{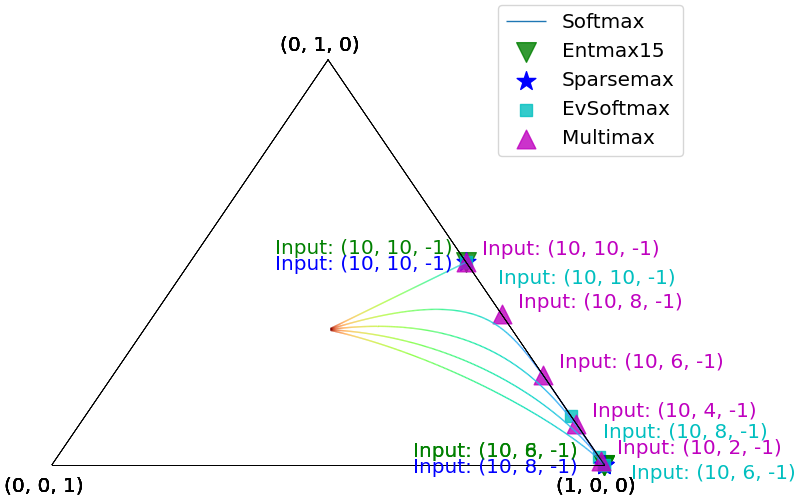}
    \caption{SoftMax output depends on the temperature, which we show by the color coding from dark blue (low temperature) to red (high temperature). Sparse SoftMax variants collapse multi-modality, while MultiMax successfully produces approximately \textbf{sparse} and \textbf{multi-modal} distributions.}    
        \label{fig:simplex}
    \end{subfigure}
    \hspace{0.5cm}
    \begin{subfigure}[t]{0.396\textwidth}
    \centering
    \includegraphics[trim={0 0 0 0},clip, width=0.9\textwidth]{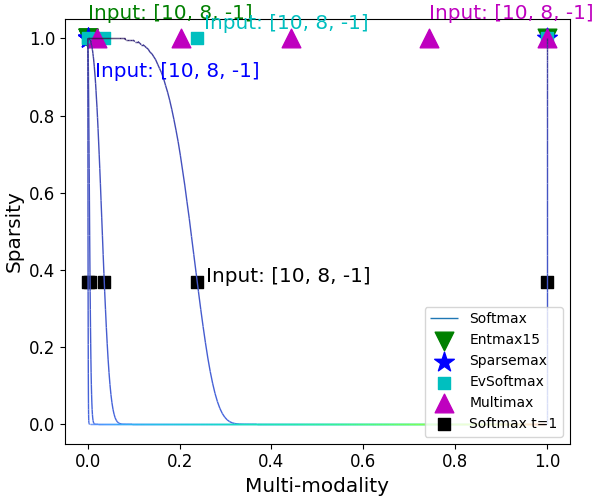}
    \caption{SoftMax and its sparse extensions are limited by the trade-off between sparsity and multi-modality, which is improved by our MultiMax. } 
       \label{fig:trade-off}
    \end{subfigure}
    
    \caption{We evaluate SoftMax, SparseMax, EntMax, EvSoftMax and MultiMax (using the parameters of a hidden layer MultiMax trained on ImageNet directly) functions on a series of example input points $\boldsymbol{v} \in \mathbb{R}^{3}$ and project the resulting distribution on a simplex $\Delta^2$. Informally, the interior of the simplex stands for trimodal distributions, the edges constitute the set of bimodal distributions, and the vertices are unimodal distributions. Notably, the above figures highlight the advantage of MultiMax's multi-modality. EntMax, Sparsemax and SoftMax with small temperature (blue colored line) yield a (quasi) uni-modal distribution, which ignore the second largest entry. In contrary, SoftMax with higher temperatures (green and orange colored line) fails to ignore the negative entry.}
    \label{fig:main}
\end{figure*}

In contrast to sparsity, multi-modality has been less discussed in the previous studies. Since attention is not supposed to be exclusive in most cases, the vanilla SoftMax, as an approximation of Argmax, does not easily comply with multi-modality. The sparse alternatives \cite{martins2016softmax, peters2019sparse, laha2018controllable} to SoftMax have even a larger tendency to not preserve the multi-modality of distributions \cite{itkina2020evidential}.

In this paper, we propose MultiMax as an alternative to SoftMax. MultiMax allows for learning when to emphasize sparsity and when to emphasize multi-modality, offering a flexible trade-off between both. At the same time, it remains piecewise differentiable such as to allow for stable gradient-based optimization. 

Specifically, MultiMax extends the traditional SoftMax by a preceding parameterized function that enables to learn distinct temperature values for particular input value ranges separately. Used within a self-attention mechanism, this facilitates for example to learn particularly low temperatures that induce sparsity for low input value ranges, i.e.~unrelated tokens can be ignored, while learning high temperatures for higher input value ranges, i.e.~several related tokens can share the attention in a multi-modal way. The improved multi-modality and sparsity brought by MultiMax is demonstrated in \cref{fig:main}. MultiMax is able to serve as a drop-in replacement of SoftMax in any applications and adapt to an appropriate form via training.

After a theoretic analysis, we show empirically that MultiMax can improve the attention mechanism and is an effective classifier output activation as well. MultiMax consistently improves over SoftMax baselines in a wide range of tasks, with an increase of $0.6\%$ classification accuracy on ImageNet, an improve of $0.7$ in perplexity for language modeling on WikiText-103, and a gain of $0.3$ in BLEU score for English to German translation on WISLT-2014. 

The contributions of this paper are as follows:
\begin{itemize}
    \item We generate insights in  the trade-off between sparsity and multi-modality in SoftMax. 
    \item We propose MultiMax -- an alternative to SoftMax with better and learnable tradeoffs between both, multi-modality and sparsity. 
    \item We show advantageous properties  of MultiMax theoretically and demonstrate  performance improvements on diverse tasks ranging from image classification over language modeling to machine translation.
\end{itemize}

\section{Related Work}
We organize the related work by first discussing related SoftMax alternatives afterwards more broadly approaches that have aimed to improve attention mechanism as well as prevent oversmoothing.

\noindent \textbf{SoftMax alternatives.}
In previous work, huge efforts have been made to pursue sparsity. Sparsemax \cite{martins2016softmax} and its generalization EntMax-$\alpha$ \cite{peters2019sparse} are sparse SoftMax variants through thresholding the output probability. Although the hyperparameter $\alpha$ is supposed to control the degree of sparsity, the functions lack full support for $\alpha>1$. Another variant, in principle similar to EntMax-1.5, with control of the sparsity is Sparsehourglass \cite{laha2018controllable}. As output activation of a classifier, these approaches require alternative losses to enable gradient-based optimization. Yet, this can cause slow convergence and training instability as well as an additional approximation error.  
Ev-SoftMax \cite{chen2021evidential} additionally
reveals that these sparse SoftMax variants could harm multi-modality. It achieves sparsification by zeroing out input entries smaller than average and provides a training-time modification strategy to enable gradient-based training. This is indeed similar to the broadly adopted top-k selection of SoftMax output, e.g., in attention layers of vision \cite{wang2022kvt, zhao2019explicit} and language \cite{gupta2021memory} transformers. 
In contrast, our MultiMax achieves sparsity and improved multi-modality at the same time without extra hyperparameters. It has also full support and thus is a drop-in replacement of SoftMax in any context. 

\noindent \textbf{Anti-oversmoothing approaches.}
Over-smoothing refers to the issue that the representations of different tokens tend to become more similar as layer depth increases. This problem is observed in both 
vision \cite{wang2022anti, gong2021improve} and language transformers \cite{shi2022revisiting}. Patch Diversification \cite{wang2022anti} combines three regularization losses to explicitly encourage diversity in patch representations. AttnScale \cite{wang2022anti} decomposes a self-attention block into low-pass and high-pass components, and rescales the high-pass component of the self-attention matrix. While these remedies have been proposed, the reason behind lacks in-depth discussion. Notably, \cite{shi2022revisiting} has attempted an analysis by relating self-attention matrix to adjacent matrix of a graph. Their claim of post-normalization being the root cause has led to further discussion, as they stick to post-normalization in the end and pre-normalization empirically performs no better than post-normalization \cite{he2020realformer}. We find that the over-smoothing problem is indeed is comparable to over-smoothing problem in GCNs \cite{chen2020measuring, oono2019graph}, and strongly related to the inevitable amount of attention assigned to irrelevant tokens. The  identity of each token degrades rapidly due to the repetitive attention operations. As shown in the studies of GCNs, sparsification \cite{rong2019dropedge, hasanzadeh2020bayesian, zheng2020robust} is a direct and effective solution.

\noindent \textbf{Attention mechanism}
A vast amount of efforts have been invested in proposing new or improving the existing attention mechanisms \cite{vaswani2017attention, velivckovic2017graph, bahdanau2014neural, gehring2016convolutional}. \cite{kim2017structured} successfully incorporated richer structural distributions into attention networks via graph encodings. \cite{niculae2017regularized} introduced a new framework for sparse and structured attention with a smoothed max operator, which can be regarded as a generalization of  softmax and sparsemax. \cite{deng2018latent} considered variational attention networks as alternatives to soft and hard attention for better learning latent variable alignment models. \cite{maruf2019selective} suggested to adopt sparse attention to selectively focus on relevant sentences in the document context for improved neural machine translation.
\cite{zhang2020sparsifying} explored the feasibility of specifying rule-based patterns to sparsify encoder outputs for improved decoding efficiency. While these approaches mainly focus on improving sparsity, our MultiMax improves both multi-modality and sparsity at the same time. Moreover, MultiMax is a universal alternative to the SoftMax function, which is not limited to the application in the attention mechanism.

\section{Background, Metrics, and Analysis}
In this section, we state the challenge of sparsity-multi-modality trade offs in reweighting functions such as softmax. Based on metrics to measure these quantities, we provide a theoretical analysis that shows the tension between those two goals in previous formulations.

\subsection{Background}
SoftMax is the most widely adopted \textbf{reweighting function} in machine learning and is formulated as follows:
\begin{definition}
    Let $\Delta^{K-1}=\{\boldsymbol{p} \in \mathbb{R}^K_{\geq0} \vert \mathbb{1}^T\boldsymbol{p}=1\}$ be the $K-1$ dimensional simplex.
SoftMax maps a vector $\boldsymbol{x} \in \mathbb{R}^{K} $ with $K \in \mathbb{Z}_+$ to a proper distribution in $\Delta ^{K-1}$: 
\begin{equation}
\label{eq:SoftMax}
    \phi_{\textit{SoftMax}}(\boldsymbol{x})_i = \dfrac{e^{tx_i}}{\sum_{k=1}^{K} e^{tx_k}},
\end{equation}
\end{definition}
where $\frac{1}{t}$ controls the entropy of the generated distribution and is often referred to as ``temperature''. 
The exponential term makes the distribution concentrated on the largest entries, which reflects the selective nature of for example the attention mechanism or multi-class classification.

\subsection{Sparsity and Multi-Modality Trade-off}
\begin{table}[t]
    \centering
    \caption{Classification accuracy on ImageNet1K using Deit-small baseline with Global Avarege Pooling (GAP) and classification token (CLS) respectively.}
    \label{tab:temperature}
    \addtolength{\tabcolsep}{-1pt}
      \scriptsize
    \begin{tabular}{@{}c|ccccccc@{}}
    \toprule
      \multirow{2}*{Model}   & \multirow{2}*{Head} &  \multicolumn{6}{c}{Temperature $\dfrac{1}{t}$}        \\
                  && 0.1 &0.5& 1 & 2 & 10 &trainable \\
         \midrule
       \multirow{2}*{Deit-small} & CLS & 5.1 & 79.9 & 79.9& \textbf{80.0} & 79.5 & 79.7 \\
        & GAP & 4.7 & 80.3 & \textbf{80.4}& 80.0 &   79.9  &80.2  \\
    \bottomrule
    \end{tabular}
\end{table}

Although sparsity seems to be easily acquired by decreasing the temperature, we find that the gain of increased sparsity comes at a cost in practice. We exemplify such an issue by 
comparing the classification performance of a transformer on ImageNet1K with different SoftMax temperatures in \cref{tab:temperature}. As shown in the table, tuning temperature is tedious and  brings no obvious advantage. Moreover, a small temperature typically provides poor learning signal and can hamper training stability, as suggested by the low accuracy for temperature $0.1$. For a better understanding of the inefficacy of temperature tuning, we follow-up with a brief theoretical study to show that the temperature tuning of SoftMax function is indeed limited by an inherent trade-off between sparsity and multi-modality.

To enable a precise analysis on the trade-off between multi-modality and sparsity, we need to define appropriate quantitative metrics for these two properties of reweighting functions.

\subsubsection{Quantifying Multi-Modality and Sparsity of Reweighting Functions}

For multi-modality and sparsity, the probabilities close to peak and zero are with no doubt the most relevant, respectively. And such relevance equivalently transfers to the largest and smallest input entries, since the studied reweighting (activation) functions should be monotonically non-decreasing \cite{ganea2019breaking, gao2017properties}. For simplification, we omit the trivial case when two entries are equal, since they remain equal after any valid function. 

 To quantitatively compare the \textbf{multi-modality} of the distributions generated by different reweighting functions $\phi$ w.r.t.~a given input $\boldsymbol{x}$, we propose the following metric  $\mathcal{M}(\boldsymbol{x})$:
\begin{definition}
\label{def:multi}
Without loss of generality, let $x_\textit{max}$ be the largest entry and $x_\textit{max} >x_n > \epsilon$, where $\epsilon$ could be any reasonable threshold for a entry to be considered relevant and N is the counts of such entries. The \textbf{Multi-Modality Metric} is given by: 
\begin{equation}
\mathcal{M}(\boldsymbol{x}) = 1 - \frac{1}{N}\sum\limits_{\epsilon<x_n<x_{\textit{max}}}^N(\phi(\boldsymbol{x})_\textit{max} - \phi(\boldsymbol{x})_n),      
\end{equation}
\end{definition}
Intuitively, this metric captures the average difference between the reweighted relevant entries  $\phi(\boldsymbol{x})_n \; \forall x_n>\epsilon$ and the maximum $\phi(\boldsymbol{x})_\textit{max}$. The average distance would be close to $0$, if all output entries are about the same (maximum multi-modality). In order to make it a large$=$better metric, we subtract it from 1.


Analogously, we build a \textbf{Sparsity Metric} for the reweighting functions upon the common $-L^1_\epsilon$ sparsity metric for vectors \cite{hurley2009comparing}, which calculates the negative sum of entries smaller than $\epsilon$. Although sparse or non-sparse is a binary status, a smooth metric is desired to additionally consider values close to zero (i.e. approximately sparse). Moreover, we would like to take the non-linear nature of such sparsity into account, i.e.,
above a reasonably small threshold, a large portion of the range from 0 to 1 is supposed to be non-sparse. In this case, a non-linear scaling (especially an approximation of a step function) helps to better reflect the actual degree of sparsity. Thus, we define the 
sparsity metric 
as follows:
\begin{definition}
\begin{equation}
    \mathcal{S}(\boldsymbol{x}) = \dfrac{1}{L}\sum\limits_{x_l<\epsilon}^L\exp{(\dfrac{s - \phi(\boldsymbol{x})_l}{s} - 1)},
\end{equation} 
\end{definition}
where $s \in [0, 1]$ can be any reference value for a non-linear scaling of the sparsity score and $L$ is the counts of entries smaller than $\epsilon$. For example, the probability of the smallest entry $x_\textit{min}$ after SoftMax ($\underset{t=1}{\text{SoftMax}}(\boldsymbol{x})_\textit{min}$) can be chosen as a reasonable reference value. Together with the exponential term, $\mathcal{S}(\boldsymbol{x})$ results in a smooth approximation of a step function, with the output range normalized to $[0, 1]$, where larger values indicate stronger degrees of sparsity.   
Having defined the two metrics, we are able to prove there exists a trade-off between them.

\subsubsection{Proofing the Trade-off}

\begin{lemma} 
\label{lemma:sparsity1}
$\mathcal{S(\boldsymbol{x})}$ is monotonically decreasing w.r.t. $\phi(\boldsymbol{x})_l$.
(See \cref{proof:1} for the proof.)
\end{lemma}
This can be easily proved by checking the partial derivative. Similar proof can be done for the following:


\begin{proposition}
\label{proposition:SoftMax}
        For a given input $\boldsymbol{x}$, the following statements hold 
         w.r.t. temperature $t$.
    \begin{enumerate}
        \item Multi-modality of SoftMax is monotonically increasing. 
        \item Sparsity of SoftMax is monotonically decreasing for $\epsilon\leq \frac{\lVert \boldsymbol{x} \rVert_1}{K}$. 
    \end{enumerate}
(See \cref{proof:SoftMax} for the proof.)
\end{proposition}

It is clear that we could increase either multi-modality or sparsity by  simply varying temperature, but at the cost of decreasing the other. As a remedy, we suggest a piece-wise modulation scheme, which modulates small and large entries via two corresponding temperatures independently. 

\section{MultiMax}
Based on our insights in the trade-off between sparsity and multi-modality in SoftMax, we propose MultiMax that reconciles those two objectives in a learnable formulation. We start by defining MultiMax that introduces two  temperature terms that control for sparsity and multi-modality respectively. We analyze improved properties that are achieved by this formulation and finally extend the concept to higher order polynomials and beyond attention mechanisms.

The following sections will provide a theorectic analysis of MultiMax, starting with its first-order form.

\subsection{First-order MultiMax}

\begin{definition}
Let $b$ and $d$ be two control parameters. We apply two corresponding temperatures $t_b$  and $t_d$ only to the entries smaller than $b$ and larger than $d$, respectively. We construct a piece-wise linear function $\sigma$ to modulate the SoftMax input $\boldsymbol{x}$, which defines the proposed MultiMax:
    \begin{equation}
  \begin{split}
   & \phi_{_\textit{MultiMax}}(\boldsymbol{x})_i = \dfrac{\exp{(\sigma(x_i))}}{\sum_{k=1}^{K} \exp{(\sigma(x_k))}},
\quad \mathrm{where}  \\ &  \resizebox{0.48\textwidth}{!}{$\sigma(x) =x + \underbrace{(1-t_{b})\textit{Max}(b-x, 0)}_{\textcolor{red}{\text{term} (1)}} + \underbrace{(t_{d}-1)\textit{Max}(x-d, 0)}_{\textcolor{red}{\text{term} (2)}}$},
 \end{split}
    \label{eq:compact_rec}
\end{equation}
We call the above function the first-order MultiMax function and we will generalize it to a higher-order version towards the end of this section. For now, the first-order MultiMax has an intuitive interpretation:
\begin{equation}
\sigma(x) =
    \begin{cases} 
               t_bx + (1-t_b)b      &  x< b\\
        x         & b \leq x \leq d \\
         t_dx + (1-t_d)d  & x > d 
   \end{cases},
   \label{eq:rec}
\end{equation}
\end{definition}
where the bias terms $(1 -t _b)b$ and $(1-t_d)d$ guarantees continuity of the modulator, e.g., $\underset{x \to b^-}{\lim} \sigma(x) = \underset{x \to b^+}{\lim} \sigma(x) = b $. To guarantee differentiability, subgradients can be defined for the turning points, e.g., $d\sigma(x)/dx=1$ at  $x=b$, please refer to \cite{boyd2003subgradient} for more details.
For $t_b>1$ and $0<t_d<1$, we could prove that MultiMax achieve a better balance between multi-modality and sparsity than SoftMax. Intuitively, a large $t_b$ pushes small entries closer to zero, while a small $t_d$ reduces the gap between large entries. Therefore, the output distribution is modulated to exhibit higher sparsity as well as multi-modality. 


To disclose the mechanism behind, we first study the impact of modulating only the small entries on the output distribution. Then we show that additionally modulating the large entries increases multi-modality further.    

\begin{figure}[ht]
    \centering
    \begin{subfigure}[t]{0.235\textwidth}
    \centering
    \includegraphics[width=\textwidth]{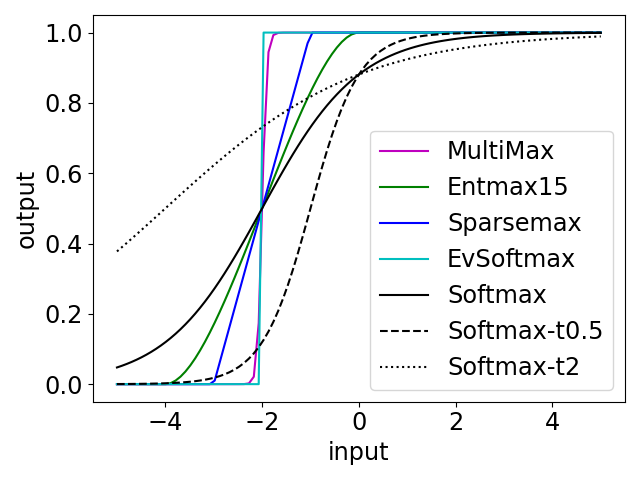}
    \caption{Input point [-2, x].}    

    \end{subfigure}
        \begin{subfigure}[t]{0.235\textwidth}
    \centering
    \includegraphics[width=\textwidth]{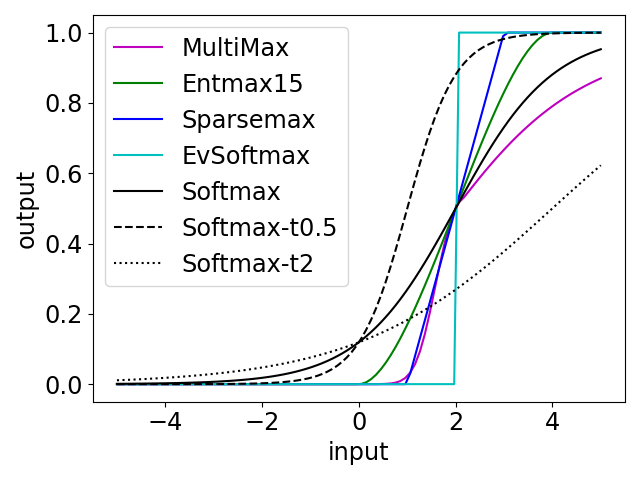}
        \caption{Input point [2, x].}          
    \end{subfigure}
    \caption{Illustration of different reweighting functions in the two-dimensional case. It can be seen clearly that MultiMax weigh the entries at small and large value ranges in a different manner, thus it does not suffer from the trade-off between sparse and multi-modal.}
    \label{fig:2d}
\end{figure}

\begin{figure}[ht]
    \centering
    \includegraphics[trim={.5cm 0 0cm 0}, width=0.42\textwidth]{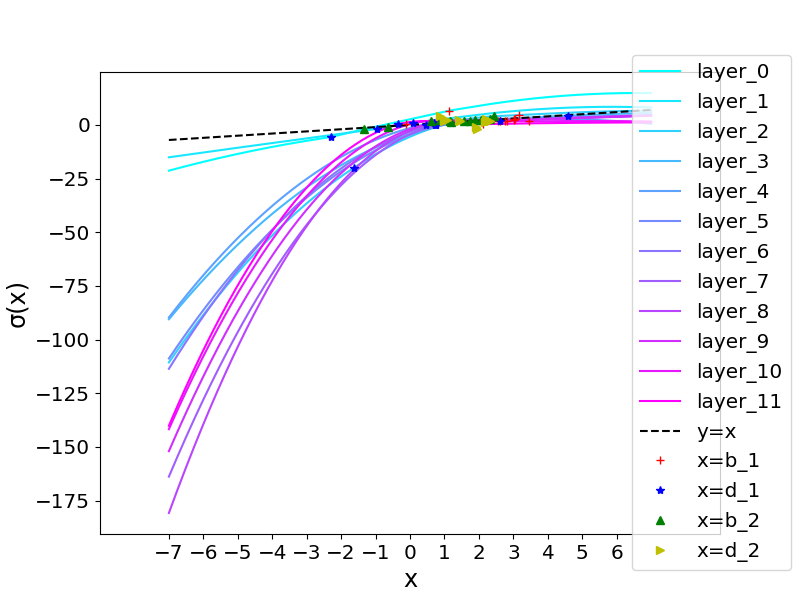}
    \caption{The learned modulator functions $\sigma$ (\cref{eq:compact_rec2}) at each layer, comparing to identity mapping of the SoftMax input $\boldsymbol{x}$ (dashed black line). All layers except for the first two converge to a form that is consistent to our analysis, i.e., low temperature (steep slope) for small entries and high temperature (flat slope) for large entries.}
    \label{fig:curve}
\end{figure}

\subsection{Improved Pareto Efficiency}
\noindent\textbf{Improving sparsity}
With the above defined metrics, we show that adding term (1) alone (denoted by MultiMax-l), i.e., modulating smaller entries, already leads to a better \textit{Pareto Optimality} \cite{buchanan1962relevance} regarding sparsity and multi-modality than SoftMax.

\begin{proposition}
\label{proposition:left}
    The following properties hold for $t_b > 1$.
    \begin{enumerate}
        \item MultiMax-l generates sparser distribution than SoftMax with temperature 1. 
        \item MultiMax-l achieves better multi-modality than SoftMax with temperature 1. 
    \end{enumerate}
    (See \cref{proof:left} for the proof.)
\end{proposition}

From the above analysis, we could see that MultiMax-l has higher \textit{Pareto Efficiency} than SoftMax: 
MultiMax-l with $t_b>1$ has both better sparsity and multi-modality than Softmax with temperature 1 (\cref{proposition:SoftMax}), and Softmax can not improve both properties at the same time by changing temperature (\cref{proposition:left}). 

\noindent \textbf{Enhancing multi-modality further}
As shown in \cref{proposition:full}, including the modulation of larger entries further enhances multi-modality while retaining better sparsity than SoftMax.

\begin{proposition}
\label{proposition:full}
    The following properties hold for $t_d < 1$ and $t_b>1$:
    \begin{enumerate}
        \item MultiMax can achieve better sparsity than SoftMax with temperature 1.
        \item MultiMax can achieve better multi-modality than MultiMax-l. 
    \end{enumerate}
    (See \cref{proof:full} for the proof.)
\end{proposition}

\subsection{Generalization}
\subsubsection{Generalization to other activations}
 Piece-wise linear activation functions are widely adopted in modern machine learning algorithms, e.g., ReLU \cite{agarap2018deep}, Leaky ReLU \cite{maas2013rectifier} and PReLU \cite{he2015delving}. Although MultiMax focuses on a different purpose, it can seen from \cref{eq:compact_rec} that the modulator/rectifier function $\sigma$ of MultiMax is a generalization of these activation functions. For example, if $b=d=0$, $t_d=1$ and $t_b=0$, then $\sigma$ is reduced to ReLU. For the rest, it can be shown easily in a similar way. 

\subsubsection{Generalization to higher-order polynomials}

So far, it has been shown that higher \textit{Pareto Efficiency} can be realized with a piece-wise linear modulation function, which belongs to the family of first-order polynomials. To obtain smoother transitions at turning points and larger capacity, second-order terms are included in our final formulation of MultiMax:
    \begin{equation}
  \begin{split}
    \resizebox{0.48\textwidth}{!}{$\sigma(x) =x + \sum\limits_{n=1}^N\underbrace{(1-t_{b_n})\textit{Max}(b_n-x, 0)^n}_{\textcolor{red}{\textit{term} (1)}} + \underbrace{(t_{d_n}-1)\textit{Max}(x-d_n, 0)^n}_{\textcolor{red}{\textit{term} (2)}}$},
 \end{split}
    \label{eq:compact_rec2}
\end{equation}
where $n$ ranges from 1 to 2. We don't include higher orders beyond the second, because it proves to be sufficient in practice.
We show in the ablation \cref{sec:ablation} that the extra nonlinearities brought by the second-order terms benefit the learning of the modulation scheme, in analogy to the previous study on activation functions \cite{hendrycks2016gaussian, clevert2015fast, elfwing2018sigmoid}.

As shown in \cref{fig:trade-off}, the output of SoftMax with varied temperatures forms a trajectory and converges to sparsemax as temperature approaches 0. EntMax-$\alpha$ stays close to the trajectory with $\alpha=1.5$, and is indeed equivalent to softmax  or SparseMax when $\alpha=1$ or $2$. MultiMax achieves, in the example, an otherwise non-reachable trade-off, with values close to the simplex that vary in two out of three possible modes. 
For a less complex illustration, we also provide the comparison with other reweighting functions with 2D inputs in \cref{fig:2d}, in which case SoftMax is equivalent to Sigmoid. While other approaches handle small and large entries equally, MultiMax provides an input-adaptive reweigthing scheme. 

We show in \cref{fig:curve} the learned modulator function of deit-small on ImageNet and compare it to the original input $\boldsymbol{x}$ (dashed black line) when used in attention layers. The learned functions at most layers (except the first two) conforms to our analysis: steeper slope for small entries (below the dashed black line on the left side means temperature smaller than 1) and flatter slope for large entries (below the dashed black line on the right side means temperature larger than 1). This conforms to our theoretical analysis that small entries should be suppressed with smaller temperature and large entries should be pushed closer with large temperature. Moreover, it is noteworthy that the need for sparsity increases as the layer goes deeper, according to the learned curves.

\subsubsection{Generalization beyond Attention}
As shown in the above analysis, the proposed MultiMax not only generalizes SoftMax, but also achieves a better Pareto optimality w.r.t.~sparsity and multi-modality with appropriate parameterization. Due to its fully parameterized formulation, it is learnable and adaptable to any scenario where a reweighting function is required. Since the need for the degree of multi-modality and sparsity may vary among different applications, we do not explicitly constrain any of the parameters and optimize them jointly with the model.

\begin{table*}[t]
        \centering
        \small
       
        \caption{Comparing to Deit \cite{touvron2021training} baseline and anti-over-smoothing methods on ImageNet-1k by replacing SoftMax with MultiMax in the attention and/or output layers. * denotes that results are not strictly comparable: these methods rely on a different training setup. For example, additional training epochs are adopted by both works,  talking-head \cite{shazeer2020talking} and a higher drop-path \cite{huang2016deep} rate are applied together with Patch Diversification.  
    }
    \label{tab:deit}
        \vspace{0.2cm}
        \scriptsize
         \begin{tabular}{@{}lccccccc@{}}
           \toprule
            \multirow{2}*{Model} & \multirow{2}*{Method}   & \multirow{2}*{Parameters} & \multirow{2}*{Epochs}  &  \multicolumn{2}{c}{Modulation}   & \multirow{2}*{Acc. (\%)}\\
           &  && &Output & Attention& &\\
            \midrule
            \multirow{2}*{Deit-tiny} & SoftMax & \multirow{2}*{5M} &300 &N/A & N/A &    72.8 \\
              &   MultiMax   & &300 & \cmark&\cmark& \textbf{73.4} \\
              \hline
            \multirow{4}*{Deit-small}  &Softmax & \multirow{6}*{22M}& 300 &N/A & N/A &  80.4\\
            & Top-k \cite{wang2022kvt}& &300  & \cmark & N/A & 80.6  \\
             & Ev-SoftMax \cite{chen2021evidential} &&  300 &-&\cmark& 80.0 \\ 
                   & \multirow{3}*{MultiMax} & &300 & \cmark &  - &  80.7  \\
            & & &300& -& \cmark &  80.7  \\
            &  & & 300&\cmark & \cmark & \textbf{81.0}  \\ 
           \hline
              \multirow{2}*{Deit-base} & SoftMax & \multirow{2}*{86M}& 300& N/A & N/A &  82.1 \\
                 & MultiMax &&300& \cmark & \cmark & \textbf{82.6} \\
       \hline
          \multirow{4}*{Deit-small}     &   Patch Diversification \cite{gong2021vision}&  & 400 & N/A& N/A &81.2*   \\
            &AttnScale  \cite{wang2022anti} && 500 & \cmark& N/A & 80.9*  \\
            & \multirow{2}*{MultiMax} & & 400   & \cmark& \cmark&  81.2 \\
           &  &&  500 & \cmark & \cmark  &  \textbf{81.3} \\
            \bottomrule
           \end{tabular}     
\end{table*}

\subsection{Computational Efficiency}

The extra computation of MultiMax is negligible for modern machine learning algorithms:
As shown in \cref{eq:compact_rec}, the total amount of additional parameters for a 12 layer Transformer with 2nd-order MultiMax is just $8\times12=96$, because each order only contains 4 parameters, including $t_b$, $t_d$, $b$ and $d$. Moreover, the modulation function $\sigma(x)$ merely consists of cheap element-wise operations, i.e., multiplication with $t_b$ and $t_d$, subtraction with $b$ and $d$, two Max operations, addition of the two terms at each order as well as a residual addition. Thus a second-order MultiMax requires $7\times2+1=15$ extra Floating Point Operations (FLOPs) for a univariant input. For Deit-small model with input length of 256, hidden dimension of 384 and 12 layers, replacing MultiMax with SoftMax in all attention layers leads to 0.0168G extra FLOPs, i.e.~only $0.37\%$ of the original model’s 4.6G FLOPs.

In practice, customized layers often run much slower than the highly optimized built-in Pytorch layers. The performance gap between theory and practice is mainly because the PyTorch framework is eagerly evaluated and thus brings additional memory access time and kernel launch time, please refer to this page \footnote{\url{https://residentmario.github.io/pytorch-training-performance-guide/jit.html}} for more details. Thus a native Pytorch implementation of MultiMax increases the training time of Deit-small on ImageNet by about $40\%$ (0.19 s/iteration vs 0.26 s/iteration), while the increase in inference time is negligible (less than $2\%$). However, we are able to achieve a reduction from $40\%$ (native Pytorch implementation) to only about $10\%$ increase of training time (0.21 s/iteration) by implementing the $\textit{Max}$ operator with 0 as built-in ReLU function and applying torch.jit.script decorator to fuse the remaining elementwise operations of our MultiMax, following the documentation \footnote{\url{https://pytorch.org/tutorials/recipes/recipes/tuning_guide.html}}. Notably, a fully optimized implementation of MultiMax in C++ or CUDA as done with Pytorch built-in layers might further reduce the gap.

\section{Experiments}
In this section, we replace SoftMax with MultiMax in different baselines and apply them to the corresponding tasks, including image classification on ImageNet1K, langauge modeling on Wiki-Text-103 corpus and machine translation on IWSLT-2014 corpus. Experimental results demonstrate consistent improvement with MultiMax, without any extra changes, e.g.~hyperparameters or  architecture. Moreover, we provide additional insights and demonstrate that advantagesous properties, including reduced over-smoothing (\cref{sec:oversmoothing}) and improved sparsity $\&$ multi-modality (\cref{sec:stats}), are achieved.

\subsection{Benchmarking}
\subsubsection{ImageNet1K Classification}
For classification, we train the widely adopted Deit \cite{touvron2021training} from scratch on ImageNet1K as baseline. Following the same training setup, we train Deit by only replacing the SoftMax function with our MultiMax, in the attention layers and/or output layer for a fair comparison.
For training, we closely follow the training settings provided in \cite{touvron2021training} and train all the models for 300 epochs. Following the more recent works \cite{chu2021conditional, liu2021swin}, we also adopt Global Average Pooling (GAP) instead of using Class Token (CLT) as classification head. While class token causes discrepancy in attention \cite{touvron2021going} and breaks translation invariance \cite{chu2021conditional}, GAP avoids this problem and improves the accuracy. 

The results in \cref{tab:deit} show a consistent improvement by using MultiMax for both attention and output activation layers. Although those sparse SoftMax variants work well for Machine Translation tasks, most of them have issues with Deit models. Ev-SoftMax decreases the performance when used in attention layers and the training does not converge (accuracy below $10\%$) when used in the output layer. For the inferior performance of Ev-SoftMax, we hypothesize that less sparsity is required for the attention among image patches than for language tokens, and zeroing out the entries smaller than average might be too aggressive. For the unstable training, their simple training-time modification might not be sufficient. The alternative losses provided by Sparse SoftMax and EntMax-1.5 require integer labels, thus are not compatible with the widely adopted label smoothing technique in vision transformers. Training instability issues are also encountered when using SparseMax in attention layers only. Therefore, we excluded them for the image classification task.

\subsubsection{Language Modeling}
We test the effectiveness of our MultiMax further on the Language Modeling task on WikiText-103 \cite{merity2016pointer} using a 6-layer Transformer Decoder with 156M parameters. The implementation is based on the official fairseq repository\footnote{\scalebox{0.9}{\label{fairseq}\url{https://github.com/facebookresearch/fairseq}}}
and the training setup is kept as default, i.e., $5\mathrm{e}{-4}$ learning rate with a maximum of 2048 tokens per GPU for 50k iterations on 4 GPUs. The results of the baseline transformer using SoftMax attention and our MultiMax are shown in \cref{tab:lm}. We again observe a consistent improvement by applying MultiMax in the output activation for this task.

\begin{table}[ht]
    \centering
    \scriptsize
            \caption{Evaluation of the performance on WikiText-103 language modeling task by test perplexity.}
    \label{tab:lm}
    \begin{tabular}{@{}l|ccc@{}}
    \toprule
       Method  & Attention & Output &   Perplexity $\downarrow$ \\
       \midrule
       SoftMax  &   - & -& 29.4        \\
            Top-k \cite{gupta2021memory} &    \cmark & N/A &  29.1     \\
        \multirow{2}*{MultiMax} &     \cmark &  -        &   29.0       \\
         &     \cmark &  \cmark        &   \textbf{28.7}   \\
\bottomrule
    \end{tabular}
\end{table}

\begin{table}[ht]
    \centering
    \scriptsize
       \caption{Comparing to other SoftMax variants using two different baseline settings (see \cref{sec:nmt} for more details) on IWSLT 2014 English to German Translation task.}
    \label{tab:bleu2}
    \begin{tabular}{@{}c|cccc@{}}
    \toprule
        SoftMax & SparseMax& EntMax-1.5 & EvSoftMax & MultiMax\\ 
          \midrule
      34.4 ± 0.07&   28.7  ± 0.16    & 34.6±0.09  &    34.7 ± 0.06  & 34.7 ± 0.07 \\
         \bottomrule
    \end{tabular}
\end{table}

\subsubsection{Machine Translation}
\label{sec:nmt}
Following previous approaches, we also evaluate our method on the task of machine translation. We train a 38M 12-layer Transformer baseline with encoder-decoder (6 layers each) architecture \cite{vaswani2017attention} from scratch on the IWSLT2014 German to English dataset \cite{cettolo2017overview}, following the training setup provided in the fairseq repository (\cref{fairseq}). Under the same setting, we also train the transformer with our MultiMax in replacement of SoftMax in the attention layers, following the common setup in previous work. The single best checkpoint and a beam size of 5 is adopted. The detokenized SacreBLEU \cite{post2018call} scores (mean and standard deviation) of 3 runs are compared in \cref{tab:bleu2}. MultiMax performs on par with EvSoftMax and is slightly better than EntMax-1.5 for this task.

\begin{figure}[ht]
\begin{subfigure}[t]{0.25\textwidth}
    \centering
    \includegraphics[width=\textwidth]{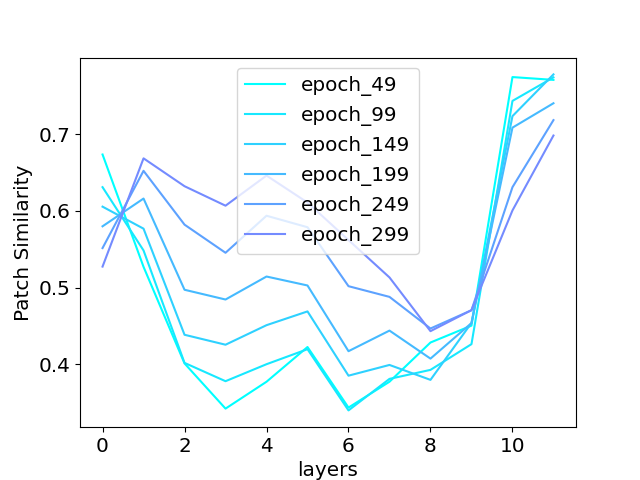}
    \caption{Softmax Deit-small}  
\end{subfigure}
\hspace{-0.5cm}
\begin{subfigure}[t]{0.25\textwidth}
    \centering
    \includegraphics[width=\textwidth]{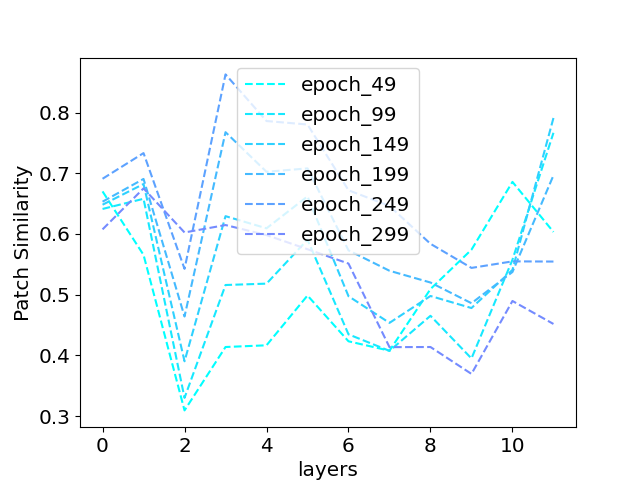}
    \caption{MultiMax Deit-small}
\end{subfigure}
\caption{Patch similarities for each layer and at different epochs. Darker color denotes the patch similarities at a larger training epoch.}
 \label{fig:similarity}   
\end{figure}

\subsection{Empirical Studies and Insights}
\label{sec:empirical}

In this section, we empirically verify the positive impact of MultiMax on the over-smoothing issue, as well as the improvement on multi-modality and sparsity in the attention scores of Deit-small trained on ImageNet1K.

\subsubsection{Analysis on Over-smoothing }
\label{sec:oversmoothing}

To validate the efficacy of our MultiMax on preventing over-smoothing, we adopt the \textit{Patch Similarity} \cite{gong2021vision} or \textit{Mean Average Distance} (MAD) \cite{chen2020measuring} metric to compare transformers using SoftMax and MultiMax on ImageNet1K. The numbers are shown 
in \cref{fig:similarity}. It can be observed that patch similarity increases as the depth grows for SoftMax attention during the entire training, whereas the patch similarity converges to a much lower level for MultiMax attention in deeper layers. We attribute this to the undesirable amount of attention assigned to irrelevant tokens which contributes the over-smoothing issue in Transformers. Moreover, it also showcases the flexibility of MultiMax's parameterized formulation, which can encourage exploration in the early stage and shift the distribution gradually towards higher sparsity as the training progresses. We have also examined the increased discrepancy between single layer attention and accumulated roll-out attention \cite{abnar2020quantifying}, which further indicates the strong connection between non-sparse SoftMax attention and the over-smoothing issue. Please refer to \cref{append:vis3} for more details.

\subsubsection{Analysis on Sparsity and Multi-modality}
\label{sec:stats}

\begin{figure}[ht]
    \centering
\includegraphics[width=0.46\textwidth, trim={0cm 0 0cm 0},]{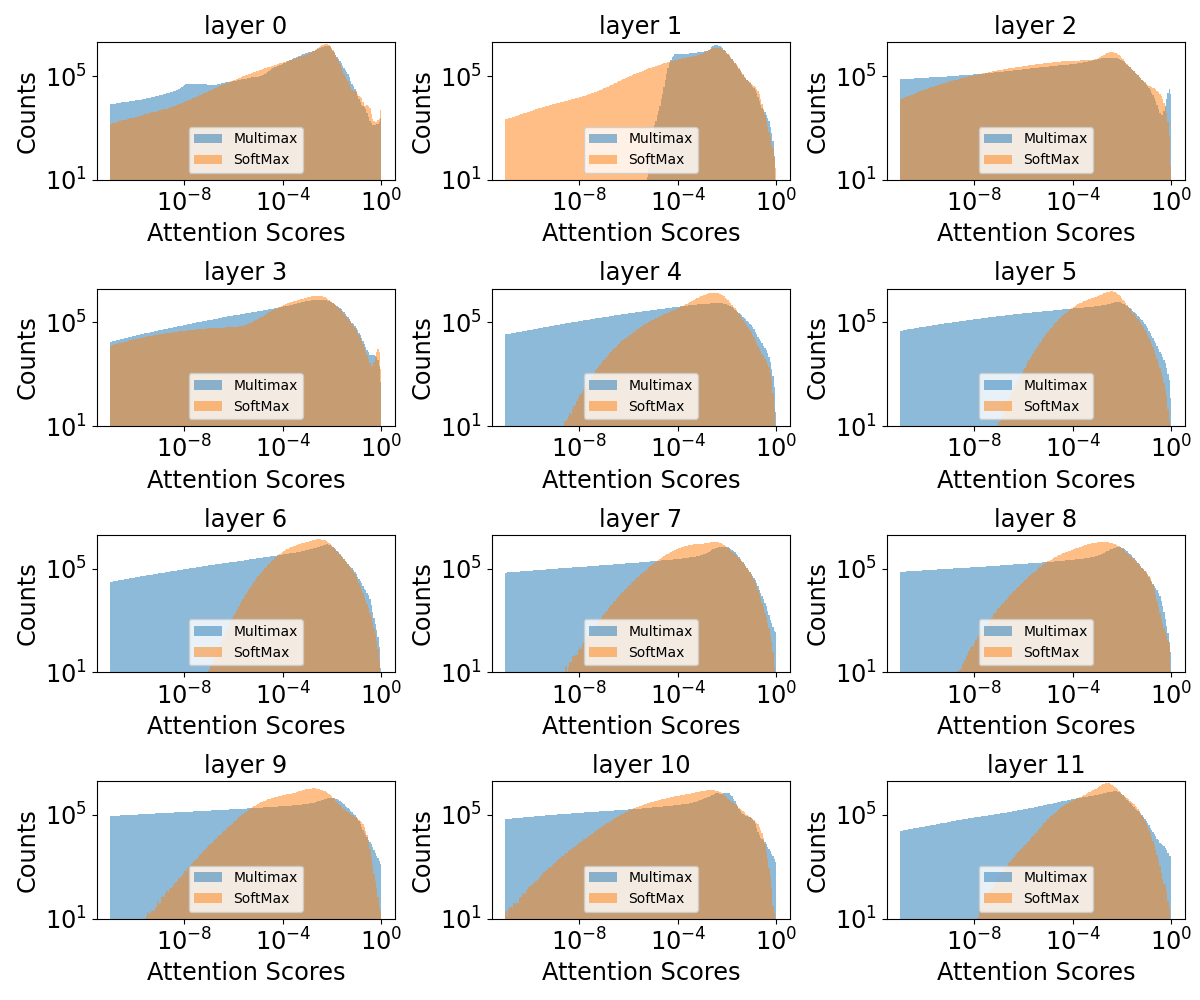}
    \caption{Histograms of the attention scores at each layer. MultiMax attention is distributed towards both ends: small scores are pushed closer to zero and more scores lie above 0.1.}
    \label{fig:hist}
\end{figure}

In this section, we empirically evaluate the impact of using our MultiMax on the sparsity of attention scores. To achieve this, we evaluate the trained model on 1000 images and collect the attention scores at each layer. 

As shown in \cref{fig:hist} in a log-log histogram, the attention scores of MultiMax are distributed more towards both ends of the score range, i.e., extremely small values near zero and large values between $0.1$ and $1$. In comparison, the attention scores of SoftMax are concentrated in the region in between, which corresponds to the bumps in the figure. Note that the number of counts are drawn at logarithmic scale, thus a small bump indeed indicates a large amount of counts. Notably, MultiMax attention behaves differently in the first two layers, which actually shows the flexibility of learning: the need for multi-modality or sparsity varies with varying context. Thus it can be a disadvantage to manually define the trade-off in advance. We also visualize the cumulative distribution of these attention scores in \cref{append:vis2}, which also indicates a stronger sparsity achieved by MultiMax.

\subsection{Ablation}
\label{sec:ablation}

To study the effect of each design component of our MultiMax independently, we conduct experiments using Deit-small as the baseline on ImageNet1K for ablation, as shown in \cref{tab:my_label}. Since the language modeling and image classification tasks are computationally heavy, we report the result of a single run with the seed unchanged for all these experiments, as commonly done for ImageNet models. 

\begin{table}[ht]
    \centering
     \scriptsize
         \caption{Impact of each MultiMax component.}
    \label{tab:my_label}
    \begin{tabular}{@{}c|cccccc@{}}
         \toprule
         Config &  term (1) & term (2) & second order & Acc \\
         \midrule
         1 &-&-&-&  80.4\\
         2 &   \cmark  &-&-&  80.6 \\
         3 & \cmark&\cmark &&  80.7 \\
         4 & \cmark & \cmark & \cmark & 81.0 \\
         \bottomrule
    \end{tabular}
\end{table}
To further validate the statistical significance of these results, we additionally conduct experiments using Deit-small with GAP on ImageNet1K and the results are recorded in \cref{tab:random}. Comparing to the relatively small standard deviation, the improvement of using MultiMax is reliable. 

\begin{table}[ht]
    \centering
        \caption{Multiple runs with random seeds using Deit-small on ImageNet1k. MultiMax shows consistent improvement over SoftMax.}
    \label{tab:random}
    \scriptsize
    \begin{tabular}{@{}c|cccccc@{}}
    \toprule
         \multirow{2}*{Method} &  \multicolumn{3}{c}{Runs} & \multirow{2}*{Mean} & \multirow{2}*{Std}\\
         & 1 & 2 & 3 & && \\
               \midrule
         SoftMax &    80.4 & 80.3 & 80.3   &  80.3 & 0.05\\
         MultiMax &   \bf{81.0}  & \bf{80.8} & \bf{80.7}   &  \bf{80.8}  & 0.12  \\
         \bottomrule
    \end{tabular}
\end{table}

\subsection{Attention Visualization}

As Transformer models \cite{vaswani2017attention, liu2021swin, zhou2022hypergraph, zhou2022sp, wang2022one}  stack a number of attention layers and aggregates the information repetitively, the attention scores at a single layer do not reflect the true information flow. To evaluate the impact on the classification more directly, we employ the well-established Grad-CAM \cite{selvaraju2017grad} to qualitatively evaluate the impact on the model's decision making. We additionally provide single layer attention scores in \cref{append:vis1} for reference.

\begin{figure}[ht]
\centering
\begin{subfigure}[t]{0.15\textwidth}
    \centering
    \includegraphics[width=\textwidth]{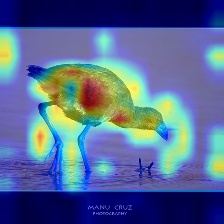} 
\end{subfigure}
\begin{subfigure}[t]{0.15\textwidth}
    \centering
    \includegraphics[width=\textwidth]{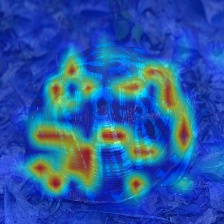}
\end{subfigure}
\begin{subfigure}[t]{0.15\textwidth}
    \centering
    \includegraphics[width=\textwidth]{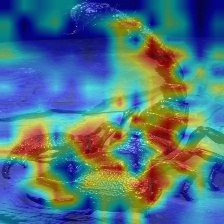}   
\end{subfigure}

\begin{subfigure}[t]{0.15\textwidth}
    \centering
    \includegraphics[width=\textwidth]{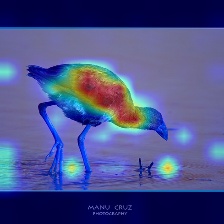}
\small 
\end{subfigure}
\begin{subfigure}[t]{0.15\textwidth}
    \centering
    \includegraphics[width=\textwidth]{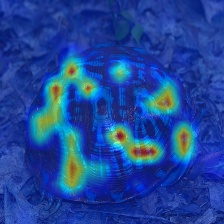}
      \small
\end{subfigure}
\begin{subfigure}[t]{0.15\textwidth}
    \centering
    \includegraphics[width=\textwidth]{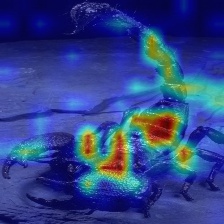}
    \small  
\end{subfigure}
    \caption{Grad-CAM of Deit-small using SoftMax (top row) and MultiMax (bottom row). 
    The MultiMax attention maps are better localized on the objects and are close to zero in most background regions, indicating sparsity at the attention level.}
    \label{fig:CAM}
\end{figure}

\section{Conclusion}
In this paper, we formalize, analyze, and evaluate the sparsity and multi-modality trade-off of SoftMax and proposed MultiMax as a remedy for tension between these two desirable objectives. Through both experimental evaluation and  analysis, we validated that MultiMax successfully learns to achieve higher multi-modality and sparsity at the same time. Although we have already demonstrated the benefits of MultiMax in attention layers and output activation of a classifier and a generative model across a wide range of tasks, we believe it has an even broader range of applications, such as in value networks and policy gradient for reinforcement learning as well as the learning of categorical distributions with Gumbel Softmax \cite{jang2016categorical}. 

\section*{Impact Statement}
This paper contributes to the understanding of core ML/AI methodology and improves the performances on a range of tasks that are broadly used as benchmark datasets in the field. Therefore, no  negative impact that would be specific to our method  is foreseeable at this point and we rather expect an overall positive impact by contributing the knowledge and understanding of these method that makes them more reliable.

\section*{Acknowledgements}
This work was partially funded by ELSA – European Lighthouse on Secure and Safe AI funded by the European Union under grant agreement No. 101070617, as well as the German Federal Ministry of Education and Research (BMBF) under the grant AIgenCY (16KIS2012). This work was supported by the Helmholtz Association's Initiative and Networking Fund on the HAICORE@FZJ partition. The authors would like to thank Dr. Stefan Kesselheim at Forschungszentrum Jülich for the kind support. 



\bibliography{example_paper}

\begin{thebibliography}{64}
\providecommand{\natexlab}[1]{#1}
\providecommand{\url}[1]{\texttt{#1}}
\expandafter\ifx\csname urlstyle\endcsname\relax
  \providecommand{\doi}[1]{doi: #1}\else
  \providecommand{\doi}{doi: \begingroup \urlstyle{rm}\Url}\fi

\bibitem[Abnar \& Zuidema(2020)Abnar and Zuidema]{abnar2020quantifying}
Abnar, S. and Zuidema, W.
\newblock Quantifying attention flow in transformers.
\newblock \emph{arXiv preprint arXiv:2005.00928}, 2020.

\bibitem[Agarap(2018)]{agarap2018deep}
Agarap, A.~F.
\newblock Deep learning using rectified linear units (relu).
\newblock \emph{arXiv preprint arXiv:1803.08375}, 2018.

\bibitem[Bahdanau et~al.(2014)Bahdanau, Cho, and Bengio]{bahdanau2014neural}
Bahdanau, D., Cho, K., and Bengio, Y.
\newblock Neural machine translation by jointly learning to align and translate.
\newblock \emph{arXiv preprint arXiv:1409.0473}, 2014.

\bibitem[Bishop \& Nasrabadi(2006)Bishop and Nasrabadi]{bishop2006pattern}
Bishop, C.~M. and Nasrabadi, N.~M.
\newblock \emph{Pattern recognition and machine learning}.
\newblock Springer, 2006.

\bibitem[Boyd et~al.(2003)Boyd, Xiao, and Mutapcic]{boyd2003subgradient}
Boyd, S., Xiao, L., and Mutapcic, A.
\newblock Subgradient methods.
\newblock \emph{lecture notes of EE392o, Stanford University, Autumn Quarter}, 2003.

\bibitem[Buchanan(1962)]{buchanan1962relevance}
Buchanan, J.~M.
\newblock The relevance of pareto optimality.
\newblock \emph{Journal of conflict resolution}, 1962.

\bibitem[Cettolo et~al.(2017)Cettolo, Federico, Bentivogli, Niehues, St{\"u}ker, Sudoh, Yoshino, and Federmann]{cettolo2017overview}
Cettolo, M., Federico, M., Bentivogli, L., Niehues, J., St{\"u}ker, S., Sudoh, K., Yoshino, K., and Federmann, C.
\newblock Overview of the iwslt 2017 evaluation campaign.
\newblock In \emph{Proceedings of the 14th International Workshop on Spoken Language Translation}, 2017.

\bibitem[Chen et~al.(2020)Chen, Lin, Li, Li, Zhou, and Sun]{chen2020measuring}
Chen, D., Lin, Y., Li, W., Li, P., Zhou, J., and Sun, X.
\newblock Measuring and relieving the over-smoothing problem for graph neural networks from the topological view.
\newblock In \emph{Proceedings of the AAAI conference on artificial intelligence}, 2020.

\bibitem[Chen et~al.(2021)Chen, Itkina, Senanayake, and Kochenderfer]{chen2021evidential}
Chen, P., Itkina, M., Senanayake, R., and Kochenderfer, M.~J.
\newblock Evidential softmax for sparse multimodal distributions in deep generative models.
\newblock \emph{Advances in Neural Information Processing Systems (NeurIPS)}, 2021.

\bibitem[Chu et~al.(2021)Chu, Tian, Zhang, Wang, Wei, Xia, and Shen]{chu2021conditional}
Chu, X., Tian, Z., Zhang, B., Wang, X., Wei, X., Xia, H., and Shen, C.
\newblock Conditional positional encodings for vision transformers.
\newblock \emph{arXiv preprint arXiv:2102.10882}, 2021.

\bibitem[Clevert et~al.(2015)Clevert, Unterthiner, and Hochreiter]{clevert2015fast}
Clevert, D.-A., Unterthiner, T., and Hochreiter, S.
\newblock Fast and accurate deep network learning by exponential linear units (elus).
\newblock \emph{arXiv preprint arXiv:1511.07289}, 2015.

\bibitem[Deng et~al.(2018)Deng, Kim, Chiu, Guo, and Rush]{deng2018latent}
Deng, Y., Kim, Y., Chiu, J., Guo, D., and Rush, A.
\newblock Latent alignment and variational attention.
\newblock \emph{Advances in eural information processing systems (NeurIPS)}, 2018.

\bibitem[Elfwing et~al.(2018)Elfwing, Uchibe, and Doya]{elfwing2018sigmoid}
Elfwing, S., Uchibe, E., and Doya, K.
\newblock Sigmoid-weighted linear units for neural network function approximation in reinforcement learning.
\newblock \emph{Neural networks}, 2018.

\bibitem[Ganea et~al.(2019)Ganea, Gelly, B{\'e}cigneul, and Severyn]{ganea2019breaking}
Ganea, O., Gelly, S., B{\'e}cigneul, G., and Severyn, A.
\newblock Breaking the softmax bottleneck via learnable monotonic pointwise non-linearities.
\newblock \emph{International Conference on Machine Learning (ICML)}, 2019.

\bibitem[Gao \& Pavel(2017)Gao and Pavel]{gao2017properties}
Gao, B. and Pavel, L.
\newblock On the properties of the softmax function with application in game theory and reinforcement learning.
\newblock \emph{arXiv preprint arXiv:1704.00805}, 2017.

\bibitem[Gehring et~al.(2016)Gehring, Auli, Grangier, and Dauphin]{gehring2016convolutional}
Gehring, J., Auli, M., Grangier, D., and Dauphin, Y.~N.
\newblock A convolutional encoder model for neural machine translation.
\newblock \emph{arXiv preprint arXiv:1611.02344}, 2016.

\bibitem[Gong et~al.(2021{\natexlab{a}})Gong, Wang, Li, Chandra, and Liu]{gong2021improve}
Gong, C., Wang, D., Li, M., Chandra, V., and Liu, Q.
\newblock Improve vision transformers training by suppressing over-smoothing.
\newblock \emph{arXiv preprint arXiv:2104.12753}, 2021{\natexlab{a}}.

\bibitem[Gong et~al.(2021{\natexlab{b}})Gong, Wang, Li, Chandra, and Liu]{gong2021vision}
Gong, C., Wang, D., Li, M., Chandra, V., and Liu, Q.
\newblock Vision transformers with patch diversification.
\newblock \emph{arXiv preprint arXiv:2104.12753}, 2021{\natexlab{b}}.

\bibitem[Goodfellow et~al.(2016)Goodfellow, Bengio, Courville, and Bengio]{goodfellow2016deep}
Goodfellow, I., Bengio, Y., Courville, A., and Bengio, Y.
\newblock \emph{Deep learning}.
\newblock MIT Press, 2016.

\bibitem[Gupta et~al.(2021)Gupta, Dar, Goodman, Ciprut, and Berant]{gupta2021memory}
Gupta, A., Dar, G., Goodman, S., Ciprut, D., and Berant, J.
\newblock Memory-efficient transformers via top-$ k $ attention.
\newblock \emph{arXiv preprint arXiv:2106.06899}, 2021.

\bibitem[Hasanzadeh et~al.(2020)Hasanzadeh, Hajiramezanali, Boluki, Zhou, Duffield, Narayanan, and Qian]{hasanzadeh2020bayesian}
Hasanzadeh, A., Hajiramezanali, E., Boluki, S., Zhou, M., Duffield, N., Narayanan, K., and Qian, X.
\newblock Bayesian graph neural networks with adaptive connection sampling.
\newblock \emph{International Conference on Machine Learning (ICML)}, 2020.

\bibitem[He et~al.(2015)He, Zhang, Ren, and Sun]{he2015delving}
He, K., Zhang, X., Ren, S., and Sun, J.
\newblock Delving deep into rectifiers: Surpassing human-level performance on imagenet classification.
\newblock In \emph{Proceedings of the IEEE International Conference on Computer Vision (ICCV)}, 2015.

\bibitem[He et~al.(2020)He, Ravula, Kanagal, and Ainslie]{he2020realformer}
He, R., Ravula, A., Kanagal, B., and Ainslie, J.
\newblock Realformer: Transformer likes residual attention.
\newblock \emph{arXiv preprint arXiv:2012.11747}, 2020.

\bibitem[Hendrycks \& Gimpel(2016)Hendrycks and Gimpel]{hendrycks2016gaussian}
Hendrycks, D. and Gimpel, K.
\newblock Gaussian error linear units (gelus).
\newblock \emph{arXiv preprint arXiv:1606.08415}, 2016.

\bibitem[Huang et~al.(2016)Huang, Sun, Liu, Sedra, and Weinberger]{huang2016deep}
Huang, G., Sun, Y., Liu, Z., Sedra, D., and Weinberger, K.~Q.
\newblock Deep networks with stochastic depth.
\newblock In \emph{European Conference on Computer Vision (ECCV)}, 2016.

\bibitem[Hurley \& Rickard(2009)Hurley and Rickard]{hurley2009comparing}
Hurley, N. and Rickard, S.
\newblock Comparing measures of sparsity.
\newblock \emph{IEEE Transactions on Information Theory}, 2009.

\bibitem[Itkina et~al.(2020)Itkina, Ivanovic, Senanayake, Kochenderfer, and Pavone]{itkina2020evidential}
Itkina, M., Ivanovic, B., Senanayake, R., Kochenderfer, M.~J., and Pavone, M.
\newblock Evidential sparsification of multimodal latent spaces in conditional variational autoencoders.
\newblock \emph{Advances in Neural Information Processing Systems (NeurIPS)}, 2020.

\bibitem[Jang et~al.(2016)Jang, Gu, and Poole]{jang2016categorical}
Jang, E., Gu, S., and Poole, B.
\newblock Categorical reparameterization with gumbel-softmax.
\newblock \emph{arXiv preprint arXiv:1611.01144}, 2016.

\bibitem[Jia \& Liang(2017)Jia and Liang]{jia2017adversarial}
Jia, R. and Liang, P.
\newblock Adversarial examples for evaluating reading comprehension systems.
\newblock \emph{arXiv preprint arXiv:1707.07328}, 2017.

\bibitem[Kim et~al.(2017)Kim, Denton, Hoang, and Rush]{kim2017structured}
Kim, Y., Denton, C., Hoang, L., and Rush, A.~M.
\newblock Structured attention networks.
\newblock \emph{International Conference on Learning Representations (ICLR)}, 2017.

\bibitem[Laha et~al.(2018)Laha, Chemmengath, Agrawal, Khapra, Sankaranarayanan, and Ramaswamy]{laha2018controllable}
Laha, A., Chemmengath, S.~A., Agrawal, P., Khapra, M., Sankaranarayanan, K., and Ramaswamy, H.~G.
\newblock On controllable sparse alternatives to softmax.
\newblock \emph{Advances in Neural Information Processing Systems (NeurIPS)}, 2018.

\bibitem[LeCun et~al.(2015)LeCun, Bengio, and Hinton]{lecun2015deep}
LeCun, Y., Bengio, Y., and Hinton, G.
\newblock Deep learning.
\newblock \emph{nature}, 2015.

\bibitem[Liu et~al.(2021)Liu, Lin, Cao, Hu, Wei, Zhang, Lin, and Guo]{liu2021swin}
Liu, Z., Lin, Y., Cao, Y., Hu, H., Wei, Y., Zhang, Z., Lin, S., and Guo, B.
\newblock Swin transformer: Hierarchical vision transformer using shifted windows.
\newblock In \emph{Proceedings of the IEEE/CVF International Conference on Computer Vision (ICCV)}, 2021.

\bibitem[Maas et~al.(2013)Maas, Hannun, Ng, et~al.]{maas2013rectifier}
Maas, A.~L., Hannun, A.~Y., Ng, A.~Y., et~al.
\newblock Rectifier nonlinearities improve neural network acoustic models.
\newblock In \emph{International Conference on Machine Learning (ICML)}, 2013.

\bibitem[Martins \& Astudillo(2016)Martins and Astudillo]{martins2016softmax}
Martins, A. and Astudillo, R.
\newblock From softmax to sparsemax a sparse model of attention and multilabel classification.
\newblock \emph{International Conference on Machine Learning (ICML)}, 2016.

\bibitem[Maruf et~al.(2019)Maruf, Martins, and Haffari]{maruf2019selective}
Maruf, S., Martins, A.~F., and Haffari, G.
\newblock Selective attention for context-aware neural machine translation.
\newblock \emph{arXiv preprint arXiv:1903.08788}, 2019.

\bibitem[Merity et~al.(2016)Merity, Xiong, Bradbury, and Socher]{merity2016pointer}
Merity, S., Xiong, C., Bradbury, J., and Socher, R.
\newblock Pointer sentinel mixture models.
\newblock \emph{arXiv preprint arXiv:1609.07843}, 2016.

\bibitem[Niculae \& Blondel(2017)Niculae and Blondel]{niculae2017regularized}
Niculae, V. and Blondel, M.
\newblock A regularized framework for sparse and structured neural attention.
\newblock \emph{Advances in Neural Information Processing Systems (NeurIPS)}, 2017.

\bibitem[Oono \& Suzuki(2019)Oono and Suzuki]{oono2019graph}
Oono, K. and Suzuki, T.
\newblock Graph neural networks exponentially lose expressive power for node classification.
\newblock \emph{arXiv preprint arXiv:1905.10947}, 2019.

\bibitem[Peters et~al.(2019)Peters, Niculae, and Martins]{peters2019sparse}
Peters, B., Niculae, V., and Martins, A.~F.
\newblock Sparse sequence-to-sequence models.
\newblock \emph{arXiv preprint arXiv:1905.05702}, 2019.

\bibitem[Post(2018)]{post2018call}
Post, M.
\newblock A call for clarity in reporting bleu scores.
\newblock \emph{arXiv preprint arXiv:1804.08771}, 2018.

\bibitem[Rong et~al.(2019)Rong, Huang, Xu, and Huang]{rong2019dropedge}
Rong, Y., Huang, W., Xu, T., and Huang, J.
\newblock Dropedge: Towards deep graph convolutional networks on node classification.
\newblock \emph{arXiv preprint arXiv:1907.10903}, 2019.

\bibitem[Rummery \& Niranjan(1994)Rummery and Niranjan]{rummery1994line}
Rummery, G.~A. and Niranjan, M.
\newblock \emph{On-line Q-learning using connectionist systems}.
\newblock University of Cambridge, Department of Engineering Cambridge, UK, 1994.

\bibitem[Schirrmeister et~al.(2020)Schirrmeister, Zhou, Ball, and Zhang]{schirrmeister2020understanding}
Schirrmeister, R., Zhou, Y., Ball, T., and Zhang, D.
\newblock Understanding anomaly detection with deep invertible networks through hierarchies of distributions and features.
\newblock \emph{Advances in Neural Information Processing Systems (NeurIPS)}, 2020.

\bibitem[Selvaraju et~al.(2017)Selvaraju, Cogswell, Das, Vedantam, Parikh, and Batra]{selvaraju2017grad}
Selvaraju, R.~R., Cogswell, M., Das, A., Vedantam, R., Parikh, D., and Batra, D.
\newblock Grad-cam: Visual explanations from deep networks via gradient-based localization.
\newblock \emph{International Conference on Machine Learning (ICML)}, 2017.

\bibitem[Shazeer et~al.(2020)Shazeer, Lan, Cheng, Ding, and Hou]{shazeer2020talking}
Shazeer, N., Lan, Z., Cheng, Y., Ding, N., and Hou, L.
\newblock Talking-heads attention.
\newblock \emph{arXiv preprint arXiv:2003.02436}, 2020.

\bibitem[Shi et~al.(2023)Shi, Chen, Misra, Scales, Dohan, Chi, Sch{\"a}rli, and Zhou]{shi2023large}
Shi, F., Chen, X., Misra, K., Scales, N., Dohan, D., Chi, E.~H., Sch{\"a}rli, N., and Zhou, D.
\newblock Large language models can be easily distracted by irrelevant context.
\newblock 2023.

\bibitem[Shi et~al.(2022)Shi, Gao, Xu, Liang, Li, Kong, Lee, and Kwok]{shi2022revisiting}
Shi, H., Gao, J., Xu, H., Liang, X., Li, Z., Kong, L., Lee, S., and Kwok, J.~T.
\newblock Revisiting over-smoothing in bert from the perspective of graph.
\newblock \emph{arXiv preprint arXiv:2202.08625}, 2022.

\bibitem[Sutton \& Barto(2018)Sutton and Barto]{sutton2018reinforcement}
Sutton, R.~S. and Barto, A.~G.
\newblock \emph{Reinforcement learning: An introduction}.
\newblock MIT press, 2018.

\bibitem[Touvron et~al.(2021{\natexlab{a}})Touvron, Cord, Douze, Massa, Sablayrolles, and J{\'e}gou]{touvron2021training}
Touvron, H., Cord, M., Douze, M., Massa, F., Sablayrolles, A., and J{\'e}gou, H.
\newblock Training data-efficient image transformers \& distillation through attention.
\newblock \emph{International Conference on Machine Learning (ICML)}, 2021{\natexlab{a}}.

\bibitem[Touvron et~al.(2021{\natexlab{b}})Touvron, Cord, Sablayrolles, Synnaeve, and J{\'e}gou]{touvron2021going}
Touvron, H., Cord, M., Sablayrolles, A., Synnaeve, G., and J{\'e}gou, H.
\newblock Going deeper with image transformers.
\newblock In \emph{Proceedings of the IEEE/CVF International Conference on Computer Vision (ICCV)}, 2021{\natexlab{b}}.

\bibitem[Vaswani et~al.(2017)Vaswani, Shazeer, Parmar, Uszkoreit, Jones, Gomez, Kaiser, and Polosukhin]{vaswani2017attention}
Vaswani, A., Shazeer, N., Parmar, N., Uszkoreit, J., Jones, L., Gomez, A.~N., Kaiser, {\L}., and Polosukhin, I.
\newblock Attention is all you need.
\newblock \emph{Advances in Neural Information Processing Systems (NeurIPS)}, 2017.

\bibitem[Veli{\v{c}}kovi{\'c} et~al.(2017)Veli{\v{c}}kovi{\'c}, Cucurull, Casanova, Romero, Lio, and Bengio]{velivckovic2017graph}
Veli{\v{c}}kovi{\'c}, P., Cucurull, G., Casanova, A., Romero, A., Lio, P., and Bengio, Y.
\newblock Graph attention networks.
\newblock \emph{arXiv preprint arXiv:1710.10903}, 2017.

\bibitem[Wang et~al.(2022{\natexlab{a}})Wang, Du, Zhang, Li, and Zhang]{wang2022one}
Wang, H., Du, Y., Zhang, Y., Li, S., and Zhang, L.
\newblock One-stage visual relationship referring with transformers and adaptive message passing.
\newblock \emph{IEEE Transactions on Image Processing}, 2022{\natexlab{a}}.

\bibitem[Wang et~al.(2022{\natexlab{b}})Wang, Wang, Wang, Lin, Chang, Li, and Jin]{wang2022kvt}
Wang, P., Wang, X., Wang, F., Lin, M., Chang, S., Li, H., and Jin, R.
\newblock Kvt: k-nn attention for boosting vision transformers.
\newblock In \emph{European Conference on Computer Vision (ECCV)}, 2022{\natexlab{b}}.

\bibitem[Wang et~al.(2022{\natexlab{c}})Wang, Zheng, Chen, and Wang]{wang2022anti}
Wang, P., Zheng, W., Chen, T., and Wang, Z.
\newblock Anti-oversmoothing in deep vision transformers via the fourier domain analysis: From theory to practice.
\newblock \emph{arXiv preprint arXiv:2203.05962}, 2022{\natexlab{c}}.

\bibitem[Weston \& Sukhbaatar(2023)Weston and Sukhbaatar]{weston2023system}
Weston, J. and Sukhbaatar, S.
\newblock System 2 attention (is something you might need too).
\newblock \emph{arXiv preprint arXiv:2311.11829}, 2023.

\bibitem[White \& Sofge(1992)White and Sofge]{white1992role}
White, D.~A. and Sofge, D.~A.
\newblock The role of exploration in learning control.
\newblock \emph{Handbook of Intelligent Control: Neural, Fuzzy and Adaptive Approaches}, 1992.

\bibitem[Williams(1992)]{williams1992simple}
Williams, R.~J.
\newblock Simple statistical gradient-following algorithms for connectionist reinforcement learning.
\newblock \emph{Machine learning}, 1992.

\bibitem[Zhang et~al.(2020)Zhang, Titov, and Sennrich]{zhang2020sparsifying}
Zhang, B., Titov, I., and Sennrich, R.
\newblock On sparsifying encoder outputs in sequence-to-sequence models.
\newblock \emph{arXiv preprint arXiv:2004.11854}, 2020.

\bibitem[Zhao et~al.(2019)Zhao, Lin, Zhang, Ren, Su, and Sun]{zhao2019explicit}
Zhao, G., Lin, J., Zhang, Z., Ren, X., Su, Q., and Sun, X.
\newblock Explicit sparse transformer: Concentrated attention through explicit selection.
\newblock \emph{arXiv preprint arXiv:1912.11637}, 2019.

\bibitem[Zheng et~al.(2020)Zheng, Zong, Cheng, Song, Ni, Yu, Chen, and Wang]{zheng2020robust}
Zheng, C., Zong, B., Cheng, W., Song, D., Ni, J., Yu, W., Chen, H., and Wang, W.
\newblock Robust graph representation learning via neural sparsification.
\newblock \emph{International Conference on Machine Learning (ICML)}, 2020.

\bibitem[Zhou et~al.(2022{\natexlab{a}})Zhou, Cheng, Li, Fang, Geng, Xie, and Keuper]{zhou2022hypergraph}
Zhou, Y., Cheng, Z.-Q., Li, C., Fang, Y., Geng, Y., Xie, X., and Keuper, M.
\newblock Hypergraph transformer for skeleton-based action recognition.
\newblock \emph{arXiv preprint arXiv:2211.09590}, 2022{\natexlab{a}}.

\bibitem[Zhou et~al.(2022{\natexlab{b}})Zhou, Xiang, Li, Wang, Wei, Zhang, Keuper, and Hua]{zhou2022sp}
Zhou, Y., Xiang, W., Li, C., Wang, B., Wei, X., Zhang, L., Keuper, M., and Hua, X.
\newblock Sp-vit: Learning 2d spatial priors for vision transformers.
\newblock In \emph{33rd British Machine Vision Conference}, 2022{\natexlab{b}}.

\end{thebibliography}
\bibliographystyle{icml2024}

\clearpage
\appendix
\section{Lemmas}
\begin{lemma}
\label{lemma:sparsity2}
The following inequalities hold:
    \begin{align*}
    x_i &> tx_i + (1-t)b , \quad \forall x_i<b \quad \text{and} \quad \forall t>1  \\
    x_i &< tx_i + (1-t)b , \quad \forall x_i >b \quad \text{and} \quad \forall t>1 \\
    x_i &< tx_i + (1-t)b , \quad \forall x_i <b \quad \text{and} \quad \forall t<1 \\
     x_i &> tx_i + (1-t)b , \quad \forall x_i >b \quad \text{and} \quad \forall t<1 
    \end{align*}
(See \cref{proof:sparsity2} for the proof.)
\end{lemma}

\begin{lemma}
\label{lemma:hoelder}
The following inequality holds $\forall \epsilon \leq \frac{1}{L}(\sum\limits_{x_l<b}^{L}X_l -lnL)$ and $\forall t>1$: 
\begin{equation*}
    \sum\limits_{x_l<b}^{L} e^{t(x_l-x_i)} \geq \sum\limits_{x_l<b}^{L} e^{x_l-x_i} 
\end{equation*}
(See \cref{proof:hoelder} for the proof.)
\end{lemma}

\section{Proofs}

\subsection{Proof of \cref{lemma:sparsity1}}
\label{proof:1}
    $\dfrac{\partial \mathcal{S}(\boldsymbol{x})}{\partial \phi(\boldsymbol{x})_l} = -\dfrac{1}{s} \exp{(\dfrac{s - \phi(\boldsymbol{x})_l}{s} - 1)}$. $\forall s>0$ $\Rightarrow$ $-\dfrac{1}{s}<0$. Since the exponential term is always positive, we have $\dfrac{\partial \mathcal{S}(\boldsymbol{x})}{\partial \phi(\boldsymbol{x})_l}<0, \quad \forall \phi(\boldsymbol{x})_l$.

\subsection{Proof of \cref{proposition:SoftMax}}
\label{proof:SoftMax}

\begin{proof} Statement 1
\begin{align*}
\intertext{from \cref{eq:SoftMax} and \cref{def:multi}}
    \frac{\partial\mathcal{M}_(\boldsymbol{x})}{\partial{t}} &=  \frac{(x_\textit{max}-x_n)e^{\frac{x_n-x_\textit{max}}{t}}}{t^2\sum_{k=1}^{K} e^{\frac{x_k-x_\textit{max}}{t}}} \\ &+\frac{(1-e^{\frac{x_n-x_\textit{max}}{t}})\sum_{k=1}^{K} 
    \frac{x_\textit{max}-x_k}{t^2}e^{\frac{x_k-x_\textit{max}}{t}}} {(\sum_{k=1}^{K} e^{\frac{x_k-x_\textit{max}}{t}})^2}  \\
    \intertext{since $x_n-x_\textit{max}<0$, we have $0<e^{\frac{x_n-x_\textit{max}}{t}}<1$}
    \Rightarrow \quad \frac{\partial\mathcal{M}(\boldsymbol{x})}{\partial{t}} &>0 \quad \text{holds} \quad \forall t &\qedhere
\end{align*}
\end{proof}

\begin{proof} Statement 2

from \cref{eq:SoftMax} 
\begin{align*}
\frac{\partial\phi_(\boldsymbol{x})_l}{\partial t} &= \frac{\sum_{k=1}^{K} (x_k-x_l)e^{\frac{x_k-x_l}{t}}}{t^2(\sum_{k=1}^{K} e^{\frac{x_k-x_l}{t}})^2} \\
\intertext{from Chebyshev's sum inequality}
\sum_{k=1}^{K} (x_k-x_l)e^{\frac{x_k-x_l}{t}} &> \frac{1}{K}  \sum_{k=1}^{K}(x_k-x_l) \sum_{k=1}^{K} e^{\frac{x_k-x_l}{t}}\\
\intertext{since $x_l< \epsilon \leq\frac{\lVert \boldsymbol{x} \rVert_1}{K}$, we have $\sum_{k=1}^{K}(x_k-x_l)\geq0$}
\Rightarrow \quad \frac{\partial\phi_(\boldsymbol{x})_l}{\partial t}&>0 \\
\intertext{from \cref{lemma:sparsity1}}
\Rightarrow \quad \frac{\partial\mathcal{S}(\boldsymbol{x})}{\partial t} &=\frac{\partial\mathcal{S}(\boldsymbol{x})}{\partial \phi_(\boldsymbol{x})_l} \frac{\partial\phi_(\boldsymbol{x})_l}{\partial t} <0
&\qedhere
\end{align*}
\end{proof}

\subsection{Proof of \cref{lemma:sparsity2}}
\label{proof:sparsity2}
From basic laws of algebra, $x - tx - (1-t)b = 
    (1-t)(x - b)$. For $t>1$ and $x<b$, we have  $ (1-t)(x - b)> 0 \Rightarrow x> tx + (1-t)b$, and vice versa.

\subsection{Proof of \cref{lemma:hoelder}}
\label{proof:hoelder}
\begin{align*}
     \intertext{since $e^{x_l} >0$, from Hoelder's inequality, we have}
 &\sum\limits_{x_l<b}^{L} e^{x_l-x_i}= 
\sum\limits_{x_l<b}^{L} 
\left|e^{x_l-x_i}\right|^1 \cdot 1 \\&\leq
\sum\limits_{x_l<b}^{L} \left((\left|e^{x_l-x_i})\right|\right)^{t})^{\frac{1}{t}} \cdot \left(\sum\limits_{l=1}^{L}1^{\frac{t}{t-1}}\right)^{1-\frac{1}{t}} \\
\intertext{raise both sides to the power of $t$ and multiply by $L^{1-t}$}
&\Rightarrow L^{1-t}(\sum\limits_{x_l<b}^{L} e^{(x_l-x_i)})^t \leq \sum\limits_{x_l<b}^{L} e^{t(x_l-x_i)}\\
\intertext{the above inequality holds if}
&\sum\limits_{x_l<b}^{L} e^{x_l-x_i}  \leq L^{1-t}(\sum\limits_{x_l<b}^{L} e^{(x_l-x_i)})^t \\
\intertext{take the natural log on both sides}
&ln\sum\limits_{x_l<b}^{L}e^{x_l-x_i} \leq (1-t)lnL + t ln\sum\limits_{x_l<b}^{L}e^{x_l-x_i} \\
&\Rightarrow lnL \leq ln\sum\limits_{x_l<b}^{L} e^{(x_l-x_i)}\\
\intertext{since $e^x$ is convex and $x_i <\epsilon$}
&\sum\limits_{x_l<b}^{L} e^{(x_l-x_i)} \geq e ^{\sum\limits_{x_l<b}^{L} (x_l-x_i)} \geq e ^{\sum\limits_{x_l<b}^{L} (x_l-\epsilon)} \\
\intertext{the condition is satisfied for $\epsilon \leq \frac{1}{L}(\sum\limits_{x_l<b}^{L}x_l -lnL)$ \qedhere}
\end{align*} 

\subsection{Proof of \cref{proposition:left} }
When only term (1) is considered, \cref{eq:rec} is reduced to:
\begin{equation}
\sigma(x) =
    \begin{dcases} 
      t_bx + (1-t_b)b  & x < b \\
        x         & x \geq b 
   \end{dcases},
\end{equation}
and we obtain:
\begin{equation}
\begin{split}
\label{eq:recmax_left}
        &\phi_{_\textit{MultiMax-l}} (\boldsymbol{x})_i
        =   \\ &\begin{dcases} 
       \frac{e^{t_bx_l + (1-t_b)b}}{\sum_{x_l<b}^{L} e^{t_bx_l + (1-t_b)b} + \sum_{x_n\geq b}^{N} e^{x_n}} & x_l < b \\
            \frac{e^{x_n}}{\sum_{x_l<b}^{L} e^{t_bx_l + (1-t_b)b} + \sum_{x_n\geq b}^{N} e^{x_n}}    & x_l \geq b 
   \end{dcases},
\end{split}
\end{equation}
where $L$ and $N$ denote the number of entries smaller than or greater than $b$ and $L+N=K$.

\label{proof:left}
\begin{proof} Statement 1
      \begin{align*}
      \intertext{from \cref{eq:recmax_left}, $\forall x_i < \epsilon \leq b$, eliminate the numerator}
      \phi_\textit{MultiMax-l} (\boldsymbol{x})_i
        &=  
      \frac{1}{\sum\limits_{x_l<b}^{L} e^{t_b(x_l-x_i)} + \sum\limits_{x_n\geq b}^{N} e^{x_n - (t_bx_i + (1-t_b)b)}}\\
       \intertext{substitute $t_bx_i + (1-t_b)b$ with $x_i$ at lower right and $\sum\limits_{x_l<b}^{L} e^{t_b(x_l-x_i)}$ at lower left, from \cref{lemma:sparsity2} and \cref{lemma:hoelder}}
 &\leq \frac{1}{\sum\limits_{x_l<b}^{L} e^{x_l-x_i)} + \sum\limits_{x_n\geq b}^{N} e^{x_n - x_i}} \\
\Rightarrow \phi_\textit{MultiMax-l}(\boldsymbol{x})_i &< \phi_\textit{SoftMax}(\boldsymbol{x})_i &\qedhere
  \end{align*}
\end{proof} 

\begin{proof} Statement 2

Eliminate $e^{x_i}$, from \cref{eq:recmax_left}, $\forall x_i > x_j > b$
\begin{align*}
      m_\textit{MultiMax-l}
        &=   
       1- \frac{1 - e^{(x_j-x_i)}}{\sum\limits_{x_l<b}^{L} e^{t_bx_l + (1-t_b)b-x_i} + \sum\limits_{x_n\geq b}^{N} e^{(x_n - x_i)}} \\
       \intertext{substitute $(1-t_b)b - x_i$ with $-t_bx_i$, from \cref{lemma:sparsity2}}
       &>  
       1 - \frac{1- e^{(x_j-x_i)}}{\sum\limits_{x_l<b}^{L} e^{t_b(x_l-x_i)} + \sum\limits_{x_n\geq b}^{N} e^{(x_n - x_i)}} 
       \\
\intertext{substitute $\sum\limits_{x_l<b}^{L} e^{t_b(x_l-x_i)}$ with $\sum\limits_{x_l<b}^{L} e^{x_l-x_i}$, from \cref{lemma:hoelder} $\quad \forall \epsilon \leq \frac{1}{L}(\sum\limits_{x_l<b}^{L}x_l -lnL)$}
&>  1 - \frac{1- e^{(x_j-x_i)}}{\sum\limits_{x_l<b}^{L} e^{x_l-x_i} + \sum\limits_{x_n\geq b}^{N} e^{(x_n - x_i)}} 
 = \mathcal{M}_\textit{SoftMax}
 &&\qedhere
\end{align*}
\end{proof} 

\subsection{Proof of \cref{proposition:full}}

Combined \cref{eq:rec} with SoftMax, we obtain:
\begin{align}
\label{eq:recmax_full}
\begin{split}
        & \phi_{_\textit{MultiMax}}(\boldsymbol{x})_i
        = \\
        & \begin{dcases} 
       \dfrac{e^{t_bx_i+(1-t_b)b}}{\sum\limits_{x_l<b}^{L} e^{\sigma(x_l)} + \sum\limits_{b \leq x_m\leq d}^{M}e^{x_m} +\sum\limits_{x_n > d}^{N} e^{\sigma(x_n)}}    & x_i < b  \\
       \dfrac{e^{x_i}}{\sum\limits_{x_l<b}^{L} e^{\sigma(x_l)} + \sum\limits_{b \leq x_m\leq d}^{M}e^{x_m} +\sum\limits_{x_n> d}^{N} e^{\sigma(x_n)}}  & b \leq x_i \leq d \\
            \dfrac{e^{t_dx_i+(1-t_d)d}}{\sum\limits_{x_l<b}^{L} e^{\sigma(x_l)} + \sum\limits_{b \leq x_m\leq d}^{M}e^{x_m} +\sum\limits_{x_n> d}^{N} e^{\sigma(x_n)}}     & x_i > d \\
   \end{dcases},
   \end{split}
\end{align}
where $L$, $M$ and $N$ denote the number of entries belonging to different ranges and $L+M+N=K$. 

\begin{proof} Statement 1
\label{proof:full}
    \begin{align*}
        \intertext{from \cref{eq:recmax_full}, $\forall x_i <\epsilon$, eliminate the numerator,
        then substitute $x_i+(1-t_b)b$ with $t_bx_i$,
        from \cref{lemma:sparsity2}}
        &<1/(\sum\limits_{x_l<b}^{L} e^{t_b(x_l-x_i)} + \sum\limits_{b \leq x_m\leq d}^{M}e^{x_m-x_i} \\&+\sum\limits_{x_n> d}^{N} e^{t_dx_n+(1-t_d)d-t_bx_i-(1-t_b)b})\\
        \intertext{from \cref{lemma:hoelder}, if $\epsilon \leq \frac{1}{M}(\sum\limits_{x_m<b}^{M}X_m -lnM)$}
        &<1/(\sum\limits_{x_l<b}^{L} e^{x_l-x_i} + \sum\limits_{b \leq x_m\leq d}^{M}e^{x_m-x_i} \\&+\sum\limits_{x_n> d}^{N} e^{t_dx_n+(1-t_d)d-t_bx_i-(1-t_b)b})\\
        \intertext{if $\sum\limits_{x_n> d}^{N} e^{t_dx_n+(1-t_d)d-t_bx_i-(1-t_b)b} > \sum\limits_{x_n> d}^{N} e^{x_n-x_i}$}
        \Rightarrow &\phi_\textit{MultiMax}(\boldsymbol{x})_i < \phi_\textit{SoftMax}(\boldsymbol{x})_i \\
        \intertext{This is satisfied when $t_dx_n+(1-t_d)d-t_bx_i-(1-t_b)b > x_n-x_i$ holds $\forall x_n$, which can be reduced to}
        &x_i< b- \frac{1-t_d}{t_b-1}(x_n-d)\\
         \intertext{where $x_n\geq d$, $t_d<1$ and $t_b>1$, and this is satisfied for}
        &\Rightarrow \epsilon \leq b- \frac{1-t_d}{t_b-1}(x_n-d)
        & \qedhere  \\
    \end{align*}
\end{proof}

\begin{figure*}[ht]
    \centering
    \begin{subfigure}[t]{0.49\textwidth}
    \includegraphics[width=0.99\textwidth]{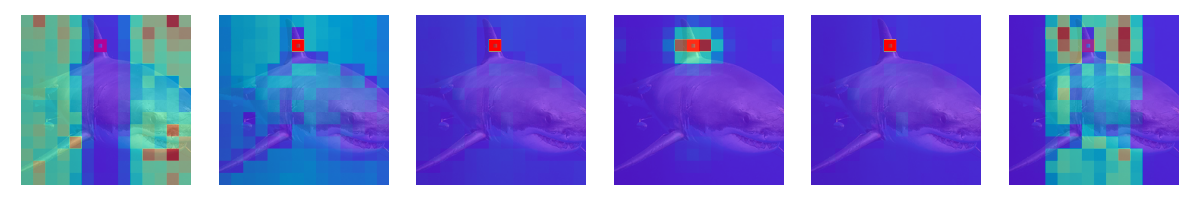}
    \end{subfigure}
    \hfill
     \begin{subfigure}[t]{0.49\textwidth}
    \includegraphics[width=0.99\textwidth]{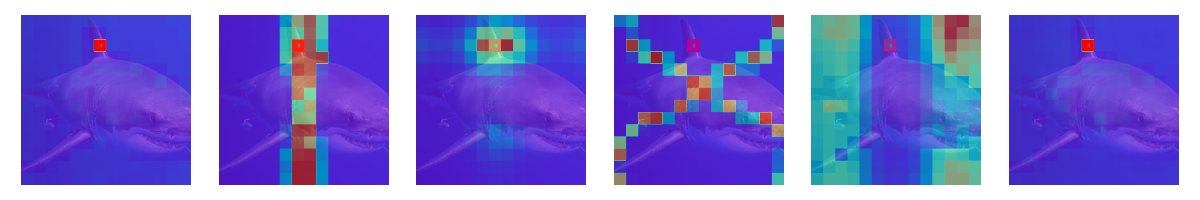}
    \end{subfigure}
        \begin{subfigure}[t]{0.49\textwidth}
    \includegraphics[width=0.99\textwidth]{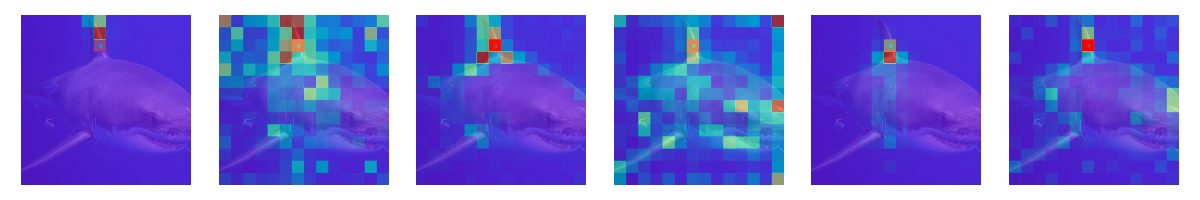}
    \end{subfigure}
    \hfill
     \begin{subfigure}[t]{0.49\textwidth}
    \includegraphics[width=0.99\textwidth]{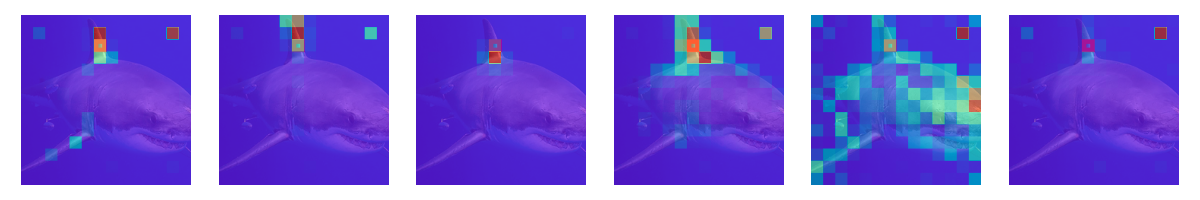}
    \end{subfigure}
        \begin{subfigure}[t]{0.49\textwidth}
    \includegraphics[width=0.99\textwidth]{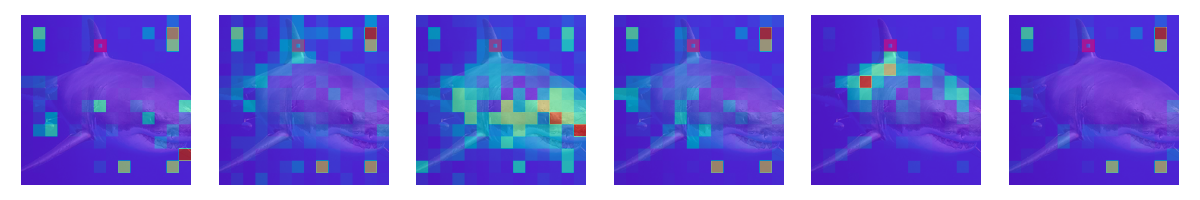}
    \end{subfigure}
    \hfill
     \begin{subfigure}[t]{0.49\textwidth}
    \includegraphics[width=0.99\textwidth]{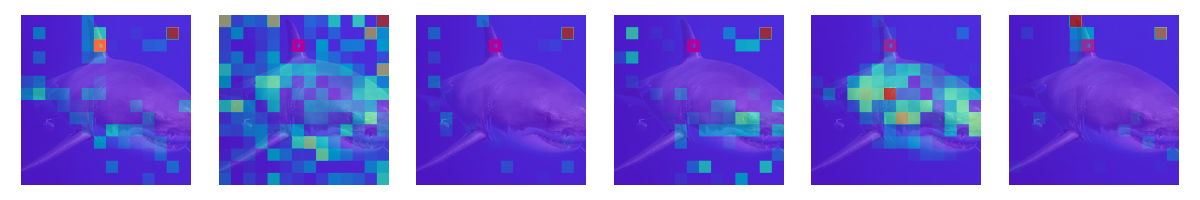}
    \end{subfigure}
    \caption{Attention scores of SoftMax (left) and MultiMax(right) at the input and hidden layers ($1^{st}$, $5^{th}$ and $10^{th}$) w.r.t query 34. The query lies on the shark fin and is marked with red square. We see, from left to right, are attention scores of 6 heads for each method, where blue refers to low attention score and red indicates a high attention score. MultiMax attention is better localized while allowing for multiple modes.}
    \label{fig:attention}
\end{figure*}

\begin{proof} Statement 2
    \begin{align*}
 \intertext{from \cref{eq:recmax_full}, $\forall x_i <\epsilon$, eliminate the numerator}
        &m_\textit{MultiMax}= 1 - (1-e^{t_d(x_j-x_i)})/ \Big( \Big. \sum\limits_{x_l<b}^{L} e^{\sigma(x_l) -t_dx_i-(1-t_d)d)} \\
        &+ \sum\limits_{b \leq x_m\leq d}^{M}e^{x_m-t_dx_i-(1-t_d)d}+ \sum\limits_{x_n> d}^{N} e^{t_d(x_n-x_i)} \Big. \Big)\\
        \intertext{since $x_j-x_i<1$ and $t_d<1,$ we have $e^{t_d(x_j-x_i)}>e^{x_j-x_i}$, also substitute $t_dx_i+(1-t_d)d$ with $t_x$, from \cref{lemma:sparsity2}}
        &> \frac{1-e^{x_j-x_i}}{\sum\limits_{x_l<b}^{L} e^{t_bx_l + (1-t_b)b -x_i} + \sum\limits_{b \leq x_m\leq d}^{M}e^{x_m-x_i}+ \sum\limits_{x_n> d}^{N} e^{t_d(x_n-x_i)}} \\
        &\Rightarrow \mathcal{M}_\textit{MultiMax}(\boldsymbol{x})>\mathcal{M}_\textit{MultiMax-l}(\boldsymbol{x}) &\qedhere
\end{align*}
\end{proof}


\section{More visualizations}
\label{append:vis}

\subsection{Single layer attention scores}
\label{append:vis1}

As mentioned in \cref{sec:empirical}, single layer attention scores are not informative for human beings, due to the complex interaction of information in deep transformer models.

\subsection{Cumulative distribution of attention scores}
\label{append:vis2}
\begin{figure}[ht]
    \centering
\includegraphics[width=0.485\textwidth, trim={0cm 0 0cm 0},]{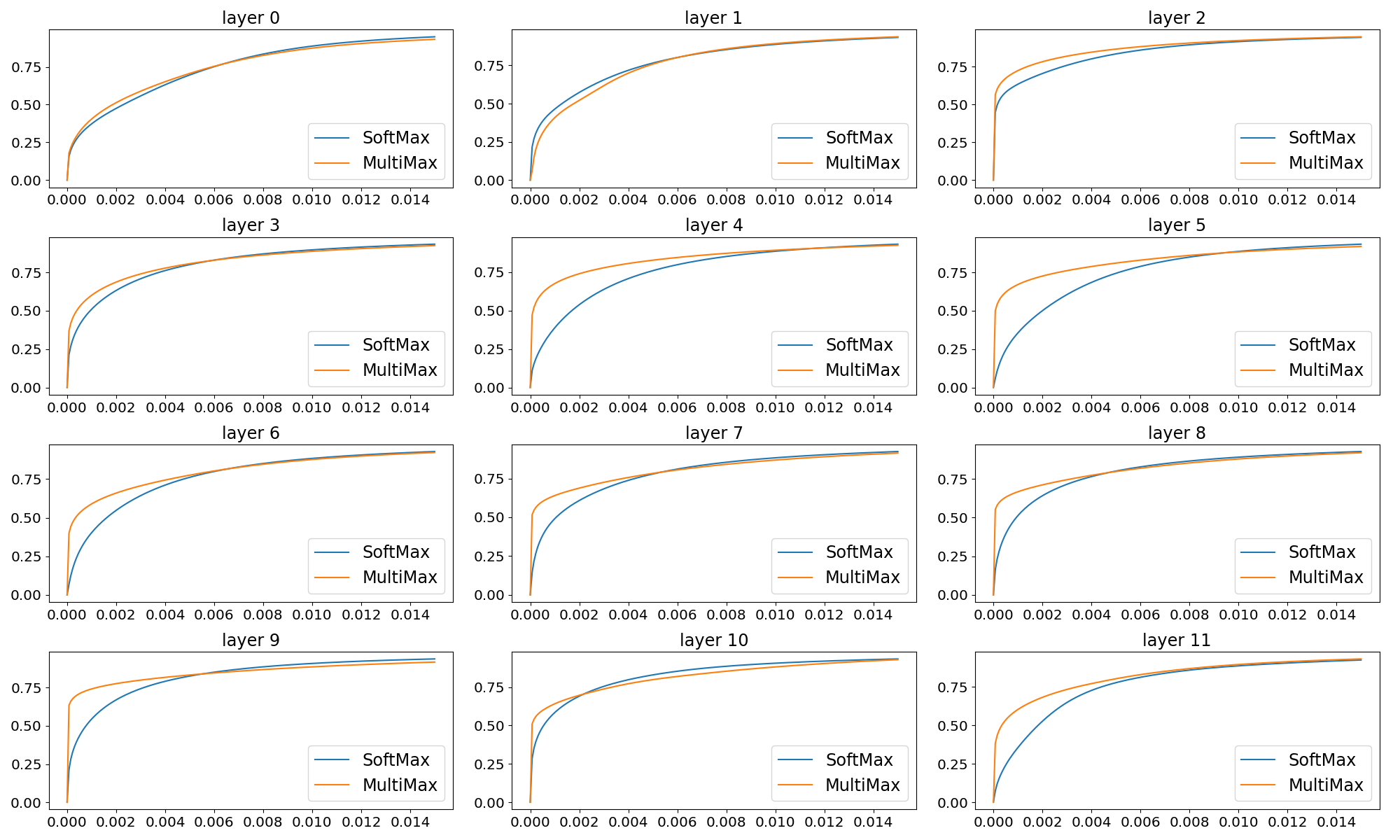}
    \caption{Cumulative distribution of the attention scores at each layer. }
    \label{fig:cumulative}
\end{figure}

We could calculate the cumulative distribution for each layer, i.e., the portion of attention scores smaller than a threshold as the thresholds increases. The result is shown in \cref{fig:cumulative}. It can be seen that for most of the layers, MultiMax results in a sparser attention distribution, i.e., a large portion of attention scores are closer to zero comparing to SoftMax attention. Notably, the first two layers' attention distributions have a smaller degree of sparsity comparing to SoftMax. This shows that a smoother distribution is desired in these two layers, as an optimized result of the training. This conforms to the observation in the previous studies that common low-level features in the shallow layers are shared across image patches \cite{schirrmeister2020understanding}. A sparse attention has a high risk of information lost.

\begin{table*}[t]
    \centering
    \scriptsize
    \begin{tabular}{c|cccccccc}
    \toprule
         Layers &	$t_{b_1}$	& $t_{d_1}$	& $t_{b_2}$	&$t_{d_2}$	& $b_1$ & $d_1$	& $b_2$	& $d_2$ \\
         \midrule
1	&1.8347933	&2.815388	&0.9864913	&0.68440557	&1.185235	&-1.208543	&-2.1076407	&1.9158255 \\
2	&1.9773115	&1.9971638	&0.985555	&0.74650276	&-0.8580209&	0.02481092	&-0.49835142	&1.9772723\\
3	&-1.1411996	&1.4711196	&1.9901285	&0.8758977	&0.18852632	&2.8039892	&2.9608543	&1.0462786 \\
4	&0.6694808	&1.206692	&1.8682657	&0.93786246	&3.4023566&	-1.5490056	&2.500237	&0.986331\\
5	&0.8902384	&1.5881691	&1.8920481	&0.72857785	&2.5070796&	-1.1942928	&1.8854694	&1.2248528 \\
6	&0.6015882	&0.87738&	2.818536	&0.96271396	&2.6490533	&0.8454426	&1.6205754	&0.89434063 \\
7	&0.8023207	&1.2427123&	3.040797	&0.84531546	&2.6984618	&1.2127148	&1.2652112	&1.2134424 \\
8	&0.64486825	&0.79173684&	2.5263662	&0.968745	&3.0230901	&0.62191963	&1.6307493&	1.6259384 \\
9	&0.5796288	&0.6852025&	3.500835&	0.99119073&	2.675157&	0.68776745	&1.3239485	&1.5808712 \\
10	&0.54873073	&0.8240905&	3.5563424&	0.9692498&	2.176066&	0.39797062	&0.9276044	&1.5223614 \\
11	&0.38645744	&0.6951747&	4.0935583&	0.9958999&	1.6583583&	0.29572898	&0.77263904	&2.9975116 \\
12	&0.16383016	&0.25565386&	3.2074118&	0.99102634&	1.6852132	&-0.04795134	&0.9796309	&2.1836245\\
\bottomrule
    \end{tabular}
    \caption{MultiMax parameters of Deit-small trained on ImageNet.}
    \label{tab:vit}
\end{table*}

\begin{table*}[t]
    \centering\scriptsize
    \begin{tabular}{c|cccccccc}
    \toprule
           Layers &	$t_{b_1}$	& $t_{d_1}$	& $t_{b_2}$	&$t_{d_2}$	& $b_1$ & $d_1$	& $b_2$	& $d_2$ \\
      \midrule
1	&0.6467285&	0.7980957&	0.98324585	&0.9649048	&0.7475586	&-0.87939453&	0.3395996&	-0.14501953\\
2	&0.69018555&	0.8063965&	0.98350525&	0.9720764	&0.25073242	&0.15991211	&0.2956543	&-0.17687988\\
3&	0.8557129&	0.79797363&	0.98939514&	0.9855194	&-0.12609863	&0.06817627&	0.14794922&	-0.14428711\\
4&	0.9662781	&0.83569336&	1.0231781&	1.0240021&	-0.07574463	&0.8510742	&-0.13220215	&0.27368164\\
5&	0.9260864	&0.9187622	&0.98670197	&1.039093	&-0.5239258	&0.51416016	&0.23999023	&0.09521484\\
6&	1.1514893&	1.152832&	0.98441315	&1.0156403&	0.1751709	&0.05374146	&-0.13269043&	-0.08825684\\
\bottomrule

    \end{tabular}
    \caption{MultiMax parameters of the 6-layer Language Transformer trained on WikiText-103.}
    \label{tab:lm_parameters}
\end{table*}

\begin{figure}[ht]
    \centering
    \includegraphics[width=0.49\textwidth]{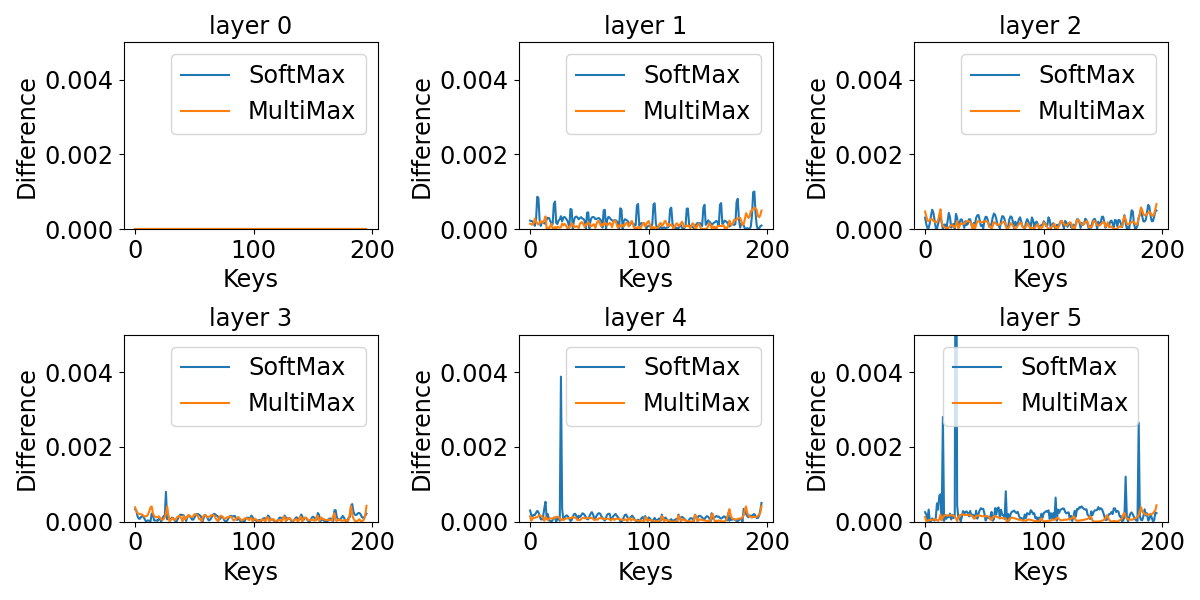}
    \caption{Comparing the discrepancy between rollout attention score and single layer attention score for SoftMax and MultiMax.}
    \label{fig:rollout_diff}
\end{figure}
\subsection{Connection between sparsification and over-smoothing}
\label{append:vis3}
As shown by \cite{abnar2020quantifying}, information originating from different input tokens gets increasingly mixed in deeper layers, and the information flow can be estimated 
by taking the attention weights out and multiplying them sequentially. Such a matrix multiplication makes the identity of each token fades exponentially, which relates to the over-smoothing problem in GCNs \cite{oono2019graph}. Considering  the information exchange across different attention heads, we take the the mean attention score over all heads out for multiplication, following the rollout technique \cite{abnar2020quantifying}. In \cref{fig:rollout_diff}, the discrepancy between the single layer and average accumulated SoftMax attention scores keeps increasing in the deeper layers. And the comparison shows a much less accumulated error for our MultiMax attention. 

\section{The learned parameters of MultiMax}

\begin{figure}[ht]
    \centering
    \includegraphics[trim={.5cm 0 0cm 0}, width=0.42\textwidth]{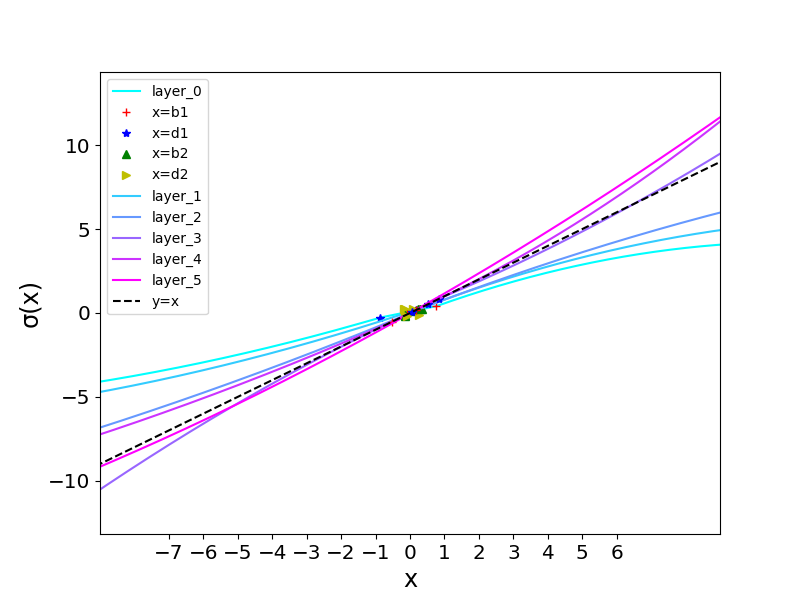}
    \caption{The learned modulator functions $\sigma$ (\cref{eq:compact_rec2}) at each layer of the 6-layer language transformer trained on WikiText-103, comparing to identity mapping of the SoftMax input $\boldsymbol{x}$ (dashed black line).}
    \label{fig:lm}
\end{figure}

In this section, we provide the learned parameters of MultiMax for reference. There are differences and similarities between the learned modulation functions of vision and language transformers, which could be observed after plotting the curves as shown in \cref{fig:lm}.:
\begin{itemize}
    \item Similarly, the need for sparsity increases as the layer goes deeper, but much less sparsity are needed in general for the language transformer compring to vision transformer, according to the learned parameters.
    \item As opposed to vision transformer, stronger multi-modality is needed at shallower layers of the language transformer.
\end{itemize}

\end{document}